\documentclass[conference]{IEEEtran} 

\usepackage{latexsym}
\usepackage{amssymb}
\usepackage{amsmath}
\usepackage{amsthm}
\usepackage{booktabs}
\usepackage{enumitem}
\usepackage{graphicx}
\usepackage{color}
\usepackage{makecell}
\usepackage{arydshln}

\newcommand{\BibTeX}{B\kern-.05em{\sc i\kern-.025em b}\kern-.08em\TeX}
\usepackage{xcolor}

\definecolor{bluecolor}{rgb}{0.1,0.5,0.8}

\usepackage[hyphens]{url}
\usepackage[breaklinks,colorlinks,allcolors=bluecolor]{hyperref}
\usepackage{comment}
\usepackage{caption}
\usepackage{subcaption}
\usepackage[capitalize]{cleveref}
\usepackage{multirow}

\title{Transferring Styles for Reduced Texture Bias and Improved Robustness in Semantic Segmentation Networks}

\author{
 \IEEEauthorblockN{1\textsuperscript{st} Ben Hamscher\textsuperscript{\textsection}}
 \IEEEauthorblockA{\textit{Department of Computer Science} \\
 \textit{Heinrich-Heine-University Düsseldorf}\\
 Düsseldorf, Germany \\
 ben.hamscher@hhu.de}
 \and
\IEEEauthorblockN{1\textsuperscript{st} Edgar Heinert\textsuperscript{\textsection}}
 \IEEEauthorblockA{\textit{Department of  Mathematics} \\
 \textit{University of Wuppertal}\\
 Wuppertal, Germany \\
 heinert@uni-wuppertal.de}
 \and
 \IEEEauthorblockN{1\textsuperscript{st} Annika Mütze\textsuperscript{\textsection}}
 \IEEEauthorblockA{\textit{Department of  Mathematics} \\
 \textit{University of Wuppertal}\\
 Wuppertal, Germany \\
 muetze@uni-wuppertal.de}
 \and[\hfill\mbox{}\par\mbox{}\hfill]
 \IEEEauthorblockN{4\textsuperscript{th} Kira Maag}
 \IEEEauthorblockA{\textit{Department of Computer Science} \\
 \textit{Heinrich-Heine-University Düsseldorf}\\
 Düsseldorf, Germany \\
 kira.maag@hhu.de}
 \and
 \IEEEauthorblockN{5\textsuperscript{th} Matthias Rottmann}
 \IEEEauthorblockA{\textit{Department of  Mathematics} \\
 \textit{University of Wuppertal}\\
 Wuppertal, Germany \\
 rottmann@uni-wuppertal.de}
 }

\begin{document}

\maketitle
\begingroup\renewcommand\thefootnote{\textsection}
\footnotetext{Equal contribution}
\endgroup

\begin{abstract}
Recent research has investigated the shape and texture biases of deep neural networks (DNNs) in image classification which influence their generalization capabilities and robustness. It has been shown that, in comparison to regular DNN training, training with stylized images reduces texture biases in image classification and improves robustness with respect to image corruptions. In an effort to advance this line of research, we examine whether style transfer can likewise deliver these two effects in semantic segmentation. To this end, we perform style transfer with style varying across artificial image areas. Those random areas are formed by a chosen number of Voronoi cells. The resulting style-transferred data is then used to train semantic segmentation DNNs with the objective of reducing their dependence on texture cues while enhancing their reliance on shape-based features.
In our experiments, it turns out that in semantic segmentation, style transfer augmentation reduces texture bias and strongly increases robustness with respect to common image corruptions as well as adversarial attacks. These observations hold for convolutional neural networks and transformer architectures on the Cityscapes dataset as well as on PASCAL Context, showing the generality of the proposed method.
\end{abstract}

%%%%%%%%%%%%%%%%%%%%%%%%%%%%%%%%%%%%%%%%%%%%%%%%%%%%%%%%%%%%%%%%%%%%%%%%

\section{Introduction}
\label{sec:intro}
Deep learning has revolutionized computer vision, with convolutional neural networks (CNNs)~\cite{726791,NIPS2012_c399862d} and vision transformers \cite{dosovitskiy2021imageworth16x16words, xie2021segformer} as widely established types of architectures for tasks such as image classification, object detection, and semantic segmentation. 
CNNs have achieved remarkable performance due to their ability to learn spatial hierarchies of features, effectively capturing patterns in image data. However, despite their success, these networks are inherently affected by biases that influence their generalization capabilities and robustness. One prominent bias in CNNs is the texture bias \cite{Geirhos2018ImageNettrainedCA, hermann2020origins, brendel2019approximating}, where models rely predominantly on local textural cues rather than global shape information when making predictions. This tendency contrasts sharply with human visual perception, which prioritizes shape over texture \cite{LANDAU1988299, jacob2021qualitative, Geirhos2018ImageNettrainedCA}. As a result, CNNs often struggle in tasks that require robust shape-based perception, limiting their ability to generalize beyond the distribution of their training data.

\begin{figure}[t]
    \centering
    % First row of images
    \begin{subfigure}[b]{0.24\textwidth}
        \includegraphics[trim=4cm 0cm 0cm 0cm, clip,width=\textwidth]{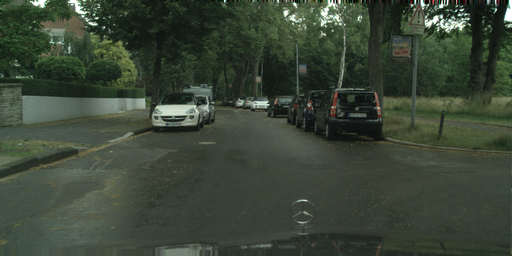}%
        %\label{fig:original}
    \end{subfigure}
    \hfill
    \begin{subfigure}[b]{0.24\textwidth}
        \includegraphics[trim=4cm 0cm 0cm 0cm, clip,width=\textwidth]{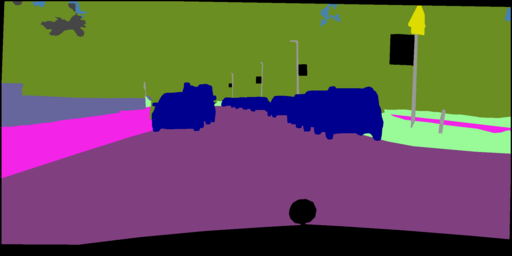}%
        %\label{fig:stylized}
    \end{subfigure}
    \\
    \vspace{3pt}
    \begin{subfigure}[b]{0.24\textwidth}
        \includegraphics[trim=4cm 0cm 0cm 0cm, clip,width=\textwidth]{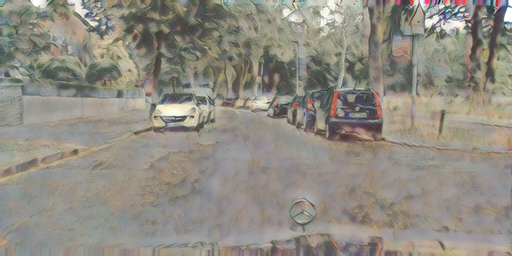}%
        %\label{fig:stylized}
    \end{subfigure}
    \hfill
    \begin{subfigure}[b]{0.24\textwidth}
        \includegraphics[trim=4cm 0cm 0cm 0cm, clip,width=\textwidth]{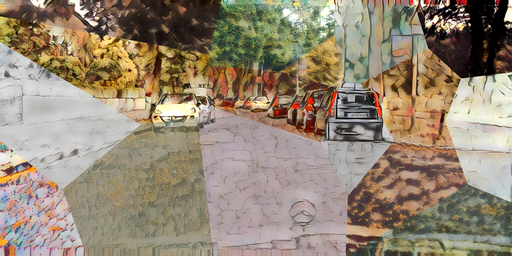}%
        %\label{fig:voronoi16}
    \end{subfigure}
    \caption{Illustration of offline AdaIN style transfer augmentation for semantic segmentation. From top left to bottom right: an original image from Cityscapes~\cite{cordts2016cityscapesdatasetsemanticurban}, the corresponding semantic segmentation mask, a full-image stylization with a single style, and a Voronoi-based stylization using 16 distinct styles.}
    \label{fig:method_illustration}
    \vspace{4pt}
\end{figure}

Recent advances in computer vision have introduced architectures beyond traditional CNNs, notably transformers, which replace convolutional operations by self-attention. Unlike CNNs, transformers process entire images holistically, allowing them to capture long-range dependencies between distant image regions. This architectural difference has been associated with a reduced texture bias \cite{NEURIPS2021_c404a5ad, zhang2022delving}, potentially leading to more shape-reliant perception or prediction. However, they can still exhibit texture bias, although to a lesser extent~\cite{10609386}.

In this work, we explore the impact of style transfer-based data augmentation in semantic segmentation. Our approach applies style transfer in a spatially localized manner using Voronoi cells (see \Cref{fig:method_illustration}) to enable the inclusion of diverse textures per image while preserving shape structure and alignment with the segmentation mask. This encourages networks to become less sensitive to abrupt texture changes and to rely more strongly on shape-based cues. While style transfer has been widely studied for artistic rendering \cite{gatys_image_2016, huang_arbitrary_2017}, its potential to reduce bias in semantic segmentation has received little attention, with most prior work focusing on classification tasks \cite{Geirhos2018ImageNettrainedCA}. One reason for this gap has been the absence of suitable metrics for texture bias in segmentation -- an issue recently addressed in \cite{heinert2025shape}, where a cue-decomposition framework for bias measurement was introduced. It uses edge-enhancing diffusion~\cite{heinert2024reducingtexturebiasdeep,Weickert2006,Weickert1994TheoreticalFO} to measure shape-dependence and Voronoi shuffling~\cite{aurenhammer1991voronoi,o1998computational} to measure texture-dependence.

We evaluate our style-transfer-based data augmentation through extensive experiments on the Cityscapes~\cite{cordts2016cityscapesdatasetsemanticurban} and the PASCAL Context~\cite{mottaghi_cvpr14} datasets using ResNet-DeepLabV3+ and SegFormer-B1 as semantic segmentation models. Both models are trained from scratch to ensure fair comparison across architectures and avoid biases from pre-training. Stylization is applied offline, with all augmented data generated prior to training using different combinations of Voronoi cell counts and proportions of stylized cells. This enables a detailed analysis of how these parameters affect shape bias, robustness, and prediction quality on the original domain. Our results show that DNNs trained on partially or fully stylized data achieve a clear reduction in texture bias with only a modest drop in original-domain performance. These models also exhibit significantly improved robustness to common corruptions~\cite{geirhos2020generalisationhumansdeepneural} and adversarial attacks~\cite{Maag_2024_WACV}. Lastly, we observe a clear trade-off between the proportion of stylized cells used for training and mIoU performance: partial stylization tends to maintain mIoU performance in the original domain while still increasing shape bias and robustness -- though the latter to a lesser extent than full stylization. Full style transfer, on the other hand, gains even more shape bias and robustness, but suffers in the original domain.

In summary, our contributions are as follows:
\begin{itemize}
    \item \textbf{Segment-Level style transfer for semantic segmentation:} We present a systematic study of neural style transfer as an offline data augmentation technique for semantic segmentation. Our approach applies style transfer locally to Voronoi-based image cells, allowing multiple distinct styles per image. This preserves shape structure and mask alignment while disrupting texture -- a combination that previously has not been explored in this context.
    
    \item \textbf{Comprehensive evaluation across architectures and Datasets:} Using stylized duplicates of Cityscapes and PASCAL Context, we rigorously analyze the effects of varying the number of Voronoi cells and the proportion of stylized cells on CNNs and transformer-based networks, including evaluations with edge-enhancing diffusion to assess shape bias.
    
    \item \textbf{Robustness gains with controllable trade-offs:} Our approach significantly reduces texture bias and boosts robustness against common corruptions and adversarial attacks. Here, the relative accuracy improves by up to $58.07$\% when performing an untargeted attack on models trained on stylized images instead of original ones and by up to $60.64$\% performing a targeted attack. 
    We further reveal that while fully stylized images yield higher shape bias and robustness, partially stylized images offer a favorable trade-off by maintaining performance on the original domain.
\end{itemize}
Our source code is publicly available at \url{https://github.com/benhamscher/transferring-styles-for-reduced-texture-bias-and-improved-robustness-in-semantic-segmentation}. 

\section{Related work}
\label{sec:rel_work}
This section reviews prior work relevant to reducing texture bias in semantic segmentation models. We begin by discussing the observed texture bias in CNNs, how it contrasts with the shape-based nature of human visual perception, and how models with stronger shape bias, such as vision transformers, have been associated with increased robustness to input corruptions and adversarial attacks. We then review neural style transfer methods (both classical and feedforward) as the foundation for our data augmentation approach. Following this, we examine how style transfer has been used to reduce texture bias in image classification, as well as recent efforts targeting texture bias in semantic segmentation.

\paragraph{Texture bias in neural networks.}
Human visual perception is predominantly shape-based, relying on the global structure of objects rather than local textural details for recognition tasks~\cite{jacob2021qualitative, LANDAU1988299}. In contrast, CNNs have been shown to exhibit a strong texture bias, favoring local patterns and fine-grained textures when making predictions. In their seminal work, \cite{Geirhos2018ImageNettrainedCA} showed that ImageNet-trained CNNs base predictions primarily on texture cues, a finding further supported by \cite{brendel2019approximating}. Recent studies on transformer-based architectures have found that these models exhibit a reduced texture bias and enhanced shape sensitivity~\cite{zhang2022delving, tuli2021convolutionalneuralnetworkstransformers,geirhos2021partial, naseer2021intriguing, tripathi2023edges}, likely due to their ability to model long-range dependencies via self-attention. While this shift contributes to improved robustness, even transformers remain more texture-biased than humans \cite{geirhos2021partial}, motivating further efforts to promote shape-based perception across architectures. Connections between an increased shape bias and robustness against image corruptions~\cite{Geirhos2018ImageNettrainedCA, heinert2025shape} and adversarial attacks~\cite{zhang2019interpretingadversarially} have been observed but \cite{mummadi2021doseenhanced} showed that training on edge maps increases shape bias without improving robustness.

\paragraph{Neural style transfer.}
Neural style transfer enables the synthesis of images that merge the content of one image with the style of another. \cite{gatys_image_2016} pioneered this approach using iterative optimization, and subsequent feedforward methods provided faster alternatives~\cite{johnson2016perceptuallossesrealtimestyle, ulyanov2016texturenetworksfeedforwardsynthesis}. Techniques such as AdaIN~\cite{huang_arbitrary_2017} and the approach by \cite{ghiasi2017exploring} enable efficient, arbitrary style transfer by aligning feature statistics. These developments form the foundation for our method which uses AdaIN-based style transfer to modify image textures while preserving the spatial alignment required for semantic segmentation. In parallel, GAN-based approaches have also been proposed for style transfer, learning to translate images across domains in a more data-driven manner. Methods like CycleGAN~\cite{zhu2017unpaired, isola2017image} enable unpaired image-to-image translation, while others such as MUNIT~\cite{huang2018multimodal} and StarGAN~\cite{choi2018stargan} support multiple styles and controllable transfer by disentangling style and content representations.

\paragraph{Style transfer for reducing texture bias in classification.}
Several studies have leveraged style transfer to mitigate texture bias in image classification. \cite{Geirhos2018ImageNettrainedCA} introduced Stylized ImageNet, a dataset created by applying AdaIN-based style transfer to ImageNet images, effectively disrupting local texture cues while preserving global shape information. This finding was supported by \cite{jackson2019style} and \cite{brendel2019approximating}, who showed that introducing diverse textures via style augmentation promotes a more human-like shape bias. \cite{kashyap2022towards} use style transfer and domain adversarial methods to mitigate texture bias. While these methods have primarily focused on classification tasks, they provide an important foundation for our work, which extends the use of style transfer to semantic segmentation. In our experiments, we employ AdaIN for stylization as it is fast, flexibly applicable, and does not require per-style training. In addition, it allows us to compare our findings to the image classification case in prior work.

\paragraph{Texture bias in semantic segmentation.}
Texture bias poses particular challenges in semantic segmentation, where maintaining the precise alignment between input images and segmentation masks is critical. \cite{kamann2020benchmarking} and \cite{hendrycks2019benchmarking} reported that segmentation models can fail under varying texture conditions even when object shapes are preserved. To better understand the contributions of different visual cues, \cite{mutze_influence_2024} decomposed images into color, shape, and texture, and \cite{Zhang2023Understanding} confirmed a strong texture bias in models such as SAM~\cite{Kirillov2023SegmentA}. The only other attempt to reduce texture bias directly in semantic segmentation models is training models on images diffused with edge-enhancing diffusion (EED) as used by \cite{heinert2024reducingtexturebiasdeep} which leads to a surprising generalization to the original image domain. Style transfer has been used to analyze properties of instance segmentation models~\cite{theodoridis2022trapped} and for debiased semantic segmentation training~\cite{li2020shape}. Recently, approaches employing neural style transfer for segmentation~\cite{kline2021improving, myeongkyun_kang__2023} have shown promising results. In contrast to these methods, for the first time, we measure the effect of style transfer training on shape bias in semantic segmentation employing the evaluation protocol by \cite{heinert2025shape} and evaluate its effect on adversarial attacks directly in semantic segmentation models. In addition, we are the first to apply style transfer at Voronoi image partition level.

\section{Stylization procedure}
\label{sec:method}
We perform offline augmentation by image stylization using adaptive instance normalization (AdaIN), originally proposed by \cite{huang_arbitrary_2017}, which enables stylization with arbitrary styles using a single stylization model. Compared to the prior work, we go beyond by partitioning images via Voronoi diagrams~\cite{aurenhammer1991voronoi,o1998computational}, stylizing Voronoi cells with different styles in order to introduce more diverse and spatially localized texture variations.  
Thereby, we encourage segmentation models to adapt to spatially (abruptly) varying textures that carry no semantic information about the ground truth. The following paragraphs provide a brief overview of the foundations of our procedure.

\paragraph{Style transfer.}
Adaptive Instance Normalization (AdaIN) was introduced by \cite{huang_arbitrary_2017} to enable efficient and flexible style transfer without requiring per-style retraining, in contrast to the computationally intensive optimization used in the original method by \cite{gatys_image_2016}. It utilizes the first four convolutional layers of an ImageNet pretrained VGG-19~\cite{simonyan2015deepconvolutionalnetworkslargescale} %
classification model as a frozen encoder $f$. For a content image $c\in \mathbb R^{C\times H \times W}$ and a style image $s\in \mathbb R^{C\times H_\text{style}\times W_\text{style}}$ an AdaIN layer aligns statistics of content and style encodings via
\begin{align}
    z =\sigma (f(s))\left( \frac{f(c)-\mu(f(c))}{\sigma(f(c))}\right)+\mu(f(s)).
\end{align}

A decoder $g$ is trained from scratch to map $z$ back to the image space, producing the stylized image $Z(c, s) = g(z) \in \mathbb{R}^{C \times H \times W}$. The decoder is supervised by a content loss,
\begin{equation}
    \mathcal{L}_c=\left\| f(g(z)) - z \right\|_2^2 \, .
\end{equation}
and a style loss based on the alignment of mean and variance across $L$ encoder layers $\phi_i$:
\begin{align}
    \mathcal{L}_s = & \sum_{i=1}^{L} \left\| \mu(\phi_i(g(z))) - \mu(\phi_i(s)) \right\|_2^2 \notag \\
    & + \left\| \sigma(\phi_i(g(z))) - \sigma(\phi_i(s)) \right\|_2^2.
\end{align}
The final loss combines both objectives to $\mathcal{L} = \mathcal{L}_c + \alpha\mathcal{L}_s$.

\paragraph{Voronoi diagrams.}
Voronoi diagrams form an established concept of geometry, with a wide range of applications~\cite{o1998computational,aurenhammer1991voronoi}. 
Given a chosen number of points \( P = \{p_1, \dots, p_n\} \subset \mathbb{R}^2 \), also called the set of sites. The Voronoi cell for a site \( p_i \) is defined as:
\begin{equation}
    V(p_i) = \{ x \in \mathbb{R}^2 : \| x - p_i \|_2 \leq \| x - p_j \|_2,\ \forall\, j \neq i \} \, ,
\end{equation}
i.e., each point in the image plane is assigned to its closest site in Euclidean distance.
Points equidistant to multiple sites form the boundaries of these cells, collectively defining the Voronoi diagram \( \mathcal{V}(P) \).

\paragraph{Voronoi diagram stylization.}
The proposed stylization method employs these concepts in a simple, yet effective way. First, images are segmented using Voronoi diagrams with a set of $n \in \mathbb{N}$ randomly chosen sites. Afterwards, each segment is stylized with a probability of $p\in[0,1]$. The parameters $n$ and $p$ can be freely adjusted based on the utilized dataset. We perform an extensive parameter study with varying quantities of Voronoi cells $n$ and stylization proportion $p$ to identify strong candidates in terms of performance on the original dataset as well as robustness.

\section{Experimental setup}\label{sec:exp_setup}
To systematically evaluate the impact of style transfer on reducing texture bias in semantic segmentation networks, we conduct comprehensive experiments using both full-image stylization and Voronoi-cell-based stylization, applied via AdaIN with random style images drawn from the Painter by Numbers dataset~\cite{painter-by-numbers}. Visual examples of the style images are provided in \Cref{fig:method_illustration} and appendix A. We assess performance across two benchmark datasets, Cityscapes~\cite{cordts2016cityscapesdatasetsemanticurban} and PASCAL Context~\cite{mottaghi_cvpr14}, using two widely adopted segmentation architectures, i.e., the CNN DeepLabV3+\cite{chen2018encoderdecoderatrousseparableconvolution} with ResNet-50 backbone~\cite{he2016deep}, and the transformer-based SegFormer MiT-B1\cite{xie2021segformer}. All models are trained from scratch using the MMSegmentation framework~\cite{mmseg2020} to ensure comparability across experiments and to isolate the effects of our data augmentation strategies from any pretraining biases. Unless stated otherwise, we adopt the standard MMSegmentation configuration for each architecture. 

\paragraph{Datasets.}
The Cityscapes~\cite{cordts2016cityscapesdatasetsemanticurban} benchmark dataset focuses on the semantic understanding of urban street scenes in Germany. It consists of $5,\!000$ images with fine annotations from $50$ cities and features $30$ classes, of which 19 are commonly used for training and evaluation~\cite{cordts2016cityscapesdatasetsemanticurban}. Of those $5,\!000$ finely annotated images, $2,\!975$ training and $500$ validation images are publicly available and used for training and later evaluations. PASCAL Context~\cite{mottaghi_cvpr14} provides additional semantic segmentation annotations for PASCAL VOC 2010 images~\cite{pascal-voc-2010}. This dataset is available in two versions of which we use the one with $33$ classes, and contains $10,\!100$ annotated training and validation images from diverse in- and outdoor scenes~\cite{mottaghi_cvpr14}. The Kaggle ``Painter by Number''~\cite{painter-by-numbers} dataset is a collection of $79,\!433$ artistic paintings by various artists. It serves as source of style images to disrupt the original texture of the two aforementioned datasets. 

\paragraph{Training details.}
For the Cityscapes dataset, we follow the standard split of $2,\!975$ training and $500$ validation images across $19$ semantic classes. DeepLabV3+ is trained on input images at a resolution of $512 \times 1,\!024$ and SegFormer at $1,\!024 \times 1,\!024$, both for $240,\!000$ iterations with a batch size of $2$.
To assess the impact of our style-transfer augmentation, we apply Voronoi-based stylization with $n \in \{4, 8, 16, 32\}$ cells per image. For the configuration with 16 cells, we further conduct an ablation study by varying the proportion of stylized image cells, $p \in \{0.25, 0.5, 0.75, 1 \}$,
to explore the trade-off between robustness and segmentation performance on the original image domain.

For the PASCAL Context dataset, we use the full set of $10,\!100$
images annotated with $33$ semantic classes, splitting it into $4,\!996$ training and $5,\!104$,
which we partition into $2,\!042$ validation and $3,\!062$ test samples to enable further parameter exploration. To preserve shape structure during stylization, we upscale the images to $1,\!440 \times 1,\!440$ for style transfer and subsequently downscale to $480 \times 480$ for training.
All experiments are conducted using the DeepLabV3+ architecture with a ResNet-50 backbone, 
trained for $160,\!000$ iterations with a batch size of $8$. 
We train on original images, on fully AdaIN-stylized images and with Voronoi-based stylization -- here we restrict experiments to $n \in \{4, 8, 16\}$ cells per image due to the lower resolution.

\paragraph{Shape bias evaluation.} To assess the shape bias of our models, we employ the recently proposed cue-decomposition-based shape bias measure (CDSB) for semantic segmentation \cite{heinert2025shape}. This method constructs two dataset variants with selectively reduced texture or shape information. 
To remove texture information while preserving the object shape, the authors use edge-enhancing diffusion (EED) \cite{heinert2024reducingtexturebiasdeep,Weickert2006,Weickert1994TheoreticalFO}, a PDE-based method that predominantly blurs color
along edges while minimizing color flow across them. We use the $\textit{EED}_\textit{mild}$ configuration, consisting of $5,\!792$ diffusion steps, a contrast parameter of $\kappa = 1/15$, a Gaussian kernel size of $\gamma=5$, and standard deviation $\sigma = \sqrt{5}$. 
Conversely, to remove shape information while preserving texture, the dataset is transformed via analogous Voronoi patch shuffling of an image and its label similarly to \cite{mutze_influence_2024}. Next, a semantic segmentation model’s mean Intersection over Union (mIoU) is computed on the EED-transformed dataset ($\textit{IoU}_{\textit{Shape}}$) and the Voronoi-patch-shuffled dataset ($\textit{IoU}_{\textit{Texture}}$). To account for dataset-specific differences in mIoU ranges, these values are normalized using constants $S$ and $T$, respectively. 
The final shape bias score is then computed as:
\begin{equation}
    \text{CDSB} = \frac{\frac{1}{S} \text{IoU}_{\text{Shape}}}{\frac{1}{S}\text{IoU}_{\text{Shape}}+\frac{1}{T}\text{IoU}_{\text{Texture}}}
\label{eq:cdsb}
\end{equation}

\begin{figure}[tb]
\centering
\setlength{\tabcolsep}{1.5pt}
\renewcommand{\arraystretch}{0.5}
\begin{tabular}{ccc}
        \includegraphics[width=0.32\columnwidth]{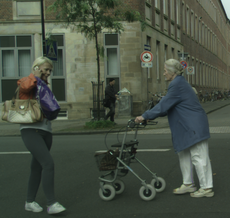}
        & \includegraphics[width=0.32\columnwidth]{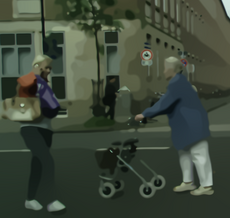}%
        & \includegraphics[width=0.32\columnwidth]{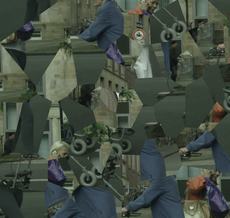}
        \\%
        Original & $\text{EED}_{\text{mild}}$ & \makecell{Voronoi-shuffled \\ (128 patches)} 
\end{tabular}
\caption[Visualization of CDSB Images]{Crop of an original Cityscapes image (left) alongside its EED-processed (center) and Voronoi patch-shuffled (right) counterparts. IoUs on the original, texture-reduced, and shape-reduced datasets form the basis of the CDSB metric. Best viewed with zoom.}
\vspace{15pt}
\label{fig:cdsb_augmentations}
\end{figure}

This score ranges from 0 to 1, where lower values indicate a stronger texture bias, and higher values indicate a stronger reliance on shape-based features. For Cityscapes, we use normalization constants $S = 25.99$ and $T = 36.43$. These values result from averaging mIoU values over a comprehensive evaluation of $22$ semantic segmentation models' mIoUs.
For PASCAL Context, we compute the constants differently as this dataset was not considered in the study of \cite{heinert2025shape}. We obtain the normalization constants $S, T$ from oracle models: $S = 38.38$ is the averaged mIoU of two DeepLabV3+ models trained on $\textit{EED}_\textit{mild}$ data with $8,\!191$ iterations, and $T = 35.11$ is the averaged mIoU of a model trained on Voronoi-shuffled images with $64$ patches. Both values are computed on the test split.
Exemplary samples of the data used to compute the different components of the shape bias score are given in \Cref{fig:cdsb_augmentations}.

\paragraph{Robustness evaluation.}
We evaluate model robustness using channel-wise corruptions applied to RGB images, adapting the methodology of \cite{geirhos2020generalisationhumansdeepneural} from grayscale to color images -– the specific corruptions procedures are also described in the appendix D. We consider contrast scaling with scaling factors in $\{0.01,0.03,0.05,0.1,0.15,0.3,0.5,1.0\}$, uniform noise randomly drawn from $[-\eta,\eta]$ with $\eta$ in $\{0.0, 0.03, 0.05, 0.1, 0.2, 0.35, 0.6, 0.9\}$ applied to both full-contrast and contrast-reduced images, low-pass filtering with Gaussian standard deviations of $1, 3, 7, 10, 15$, and $40$ pixels, and high-pass filtering with standard deviations of $0.4, 0.45, 0.55, 0.7, 1, 1.5$, and $3$ pixels.  Phase noise is applied per channel with shifts randomly drawn from $[-w, w]$, using $w \in \{0, 30, 60, 90, 120, 150, 180\}$ degrees. All experiments are conducted in triplicate and we report mean values to ensure statistical reliability.
Finally, to succinctly summarize model robustness, a \emph{Robustness Score} is calculated, defined as the ratio between the average mIoU across all corruptions and severity levels and the mIoU obtained from the original Cityscapes validation images:
\begin{equation}
    \text{Robustness Score} = \frac{\frac{1}{D}\sum_{d=1}^{D}\frac{1}{L_d}\sum_{l=1}^{L_d} \text{mIoU}_{d, l}}{\text{mIoU}_{\text{original}}},
\end{equation}
where $d=1,\ldots, D$ denotes the corruption type out of the previous list, $l = 1, \ldots, L_d$ denotes the evaluated corruptions and intensity levels from the previously stated parameters. $\text{mIoU}_{\text{original}}$ and $\text{mIoU}_{d, l}$ refer to mIoU values achieved on original validation images and images with corruption $d$ at level $l$. This score aims to summarize the results of each network on the entire set of corruptions while accounting for possibly different numbers of parameters per corruption. Normalization with mIoU on the original data makes the score a measure of relative performance under the different corruptions.

\paragraph{Baselines.} 
We establish baseline models for both Cityscapes and PASCAL Context using the same training paradigm and architectures as described before. Each model is trained three times on all variants of the Cityscapes dataset and twice on variants of PASCAL Context. Baselines are trained on three variations of each dataset: (1) The respective original dataset; (2) $\textit{EED}_\textit{mild}$, an EED version of the dataset with mild smoothing ($\kappa=1/15$, Gaussian kernel size $5$, $\sigma=\sqrt{5}$, $5,\!792$ diffusion steps)
; and (3) $\textit{EED}_\textit{strong}$, which applies stronger diffusion-based smoothing ($\kappa=1/10$, Gaussian kernel size $9$, $\sigma=3$, $5,\!792$ diffusion steps). Notably, in the CDSB evaluation, the EED baseline models serve as a form of oracle, as the mild EED variation of the dataset is used to measure shape performance.

\section{Numerical results}
\label{sec:results}

In our study, we investigate the influence of texture randomization through style transfer augmentation on four key aspects of semantic segmentation models: mIoU performance, shape bias, corruption robustness and robustness with respect to adversarial attacks.

\begin{figure*}[tb]
    \centering
    \begin{minipage}[t]{0.44\textwidth}
        \centering
        \includegraphics[width=\linewidth]{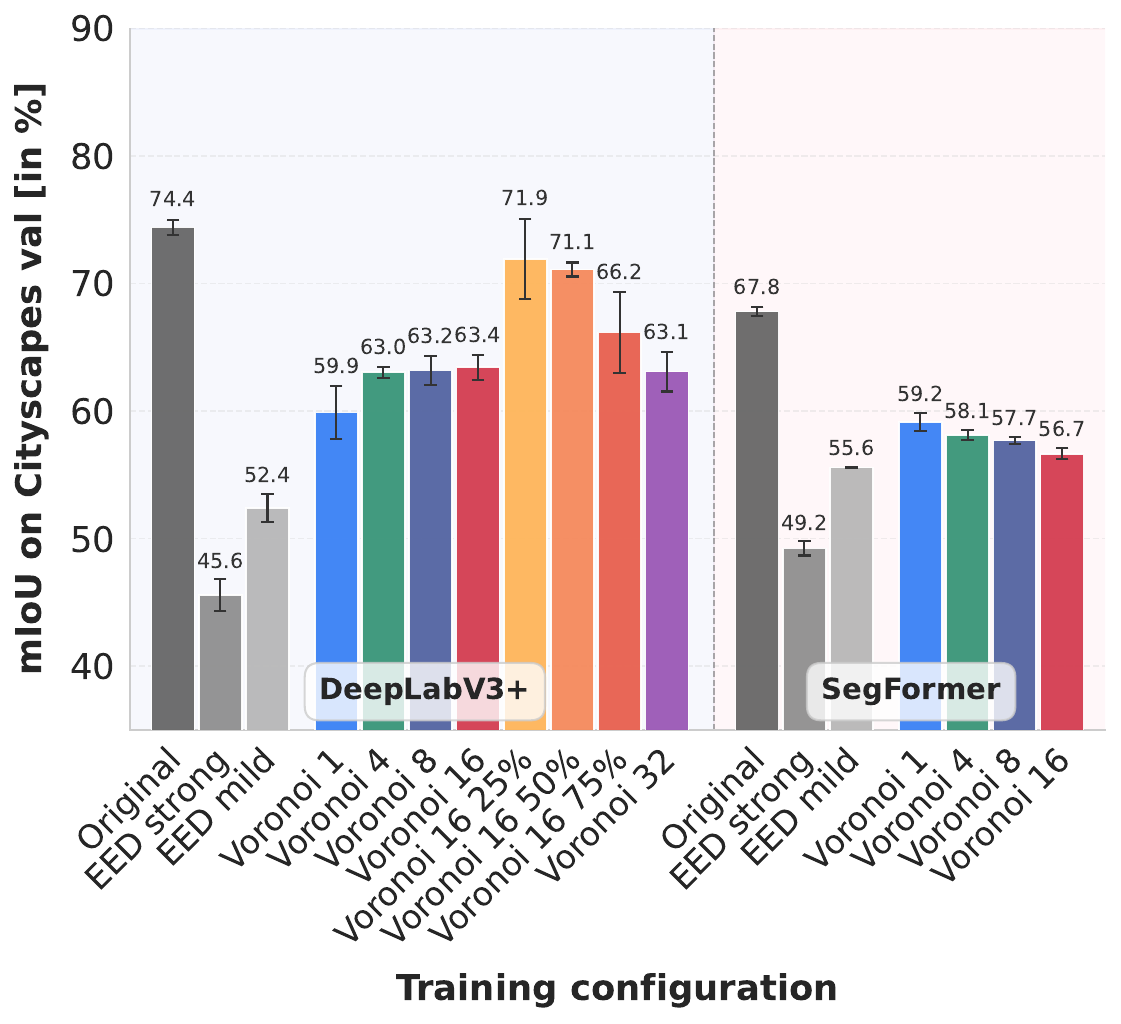}
        \caption{mIoU scores on the Cityscapes validation set for models trained with different offline augmentations. Error bars indicate mean and standard deviation of evaluations across three training runs.}
        \label{fig:cs_miou_barplot}
    \end{minipage}
    \hspace{14pt}
    \begin{minipage}[t]{0.44\textwidth}
        \centering
        \includegraphics[width=\linewidth]{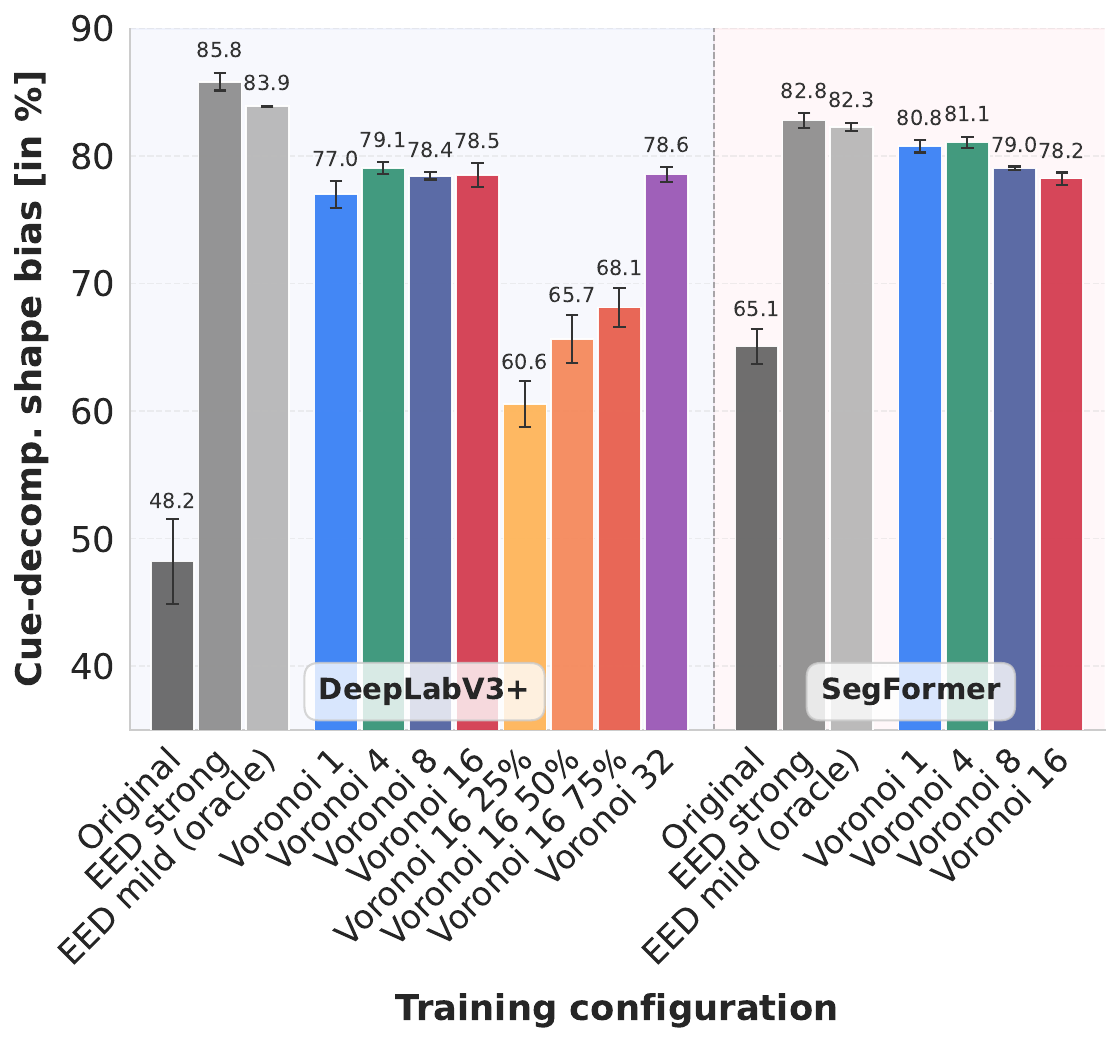}
        \caption[Comparison of Cue-Decomposition Shape Bias]{Cue-decomposition shape bias results on Citypscapes for varying offline augmentation settings. Error bars indicate mean and standard deviation of evaluations across three training runs.}
        \label{fig:cs_cdsb_barplot}
    \end{minipage}
\end{figure*}

\begin{figure}[tb]
    \centering
    \includegraphics[width=0.465\linewidth]{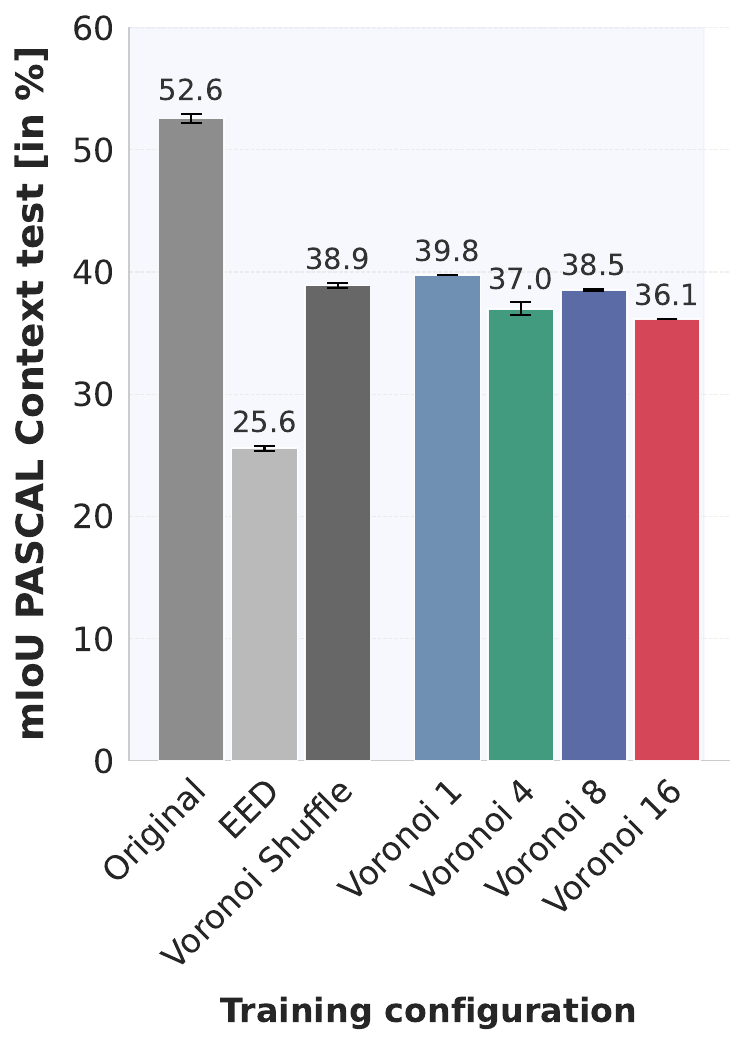}
    \includegraphics[width=0.48\linewidth]{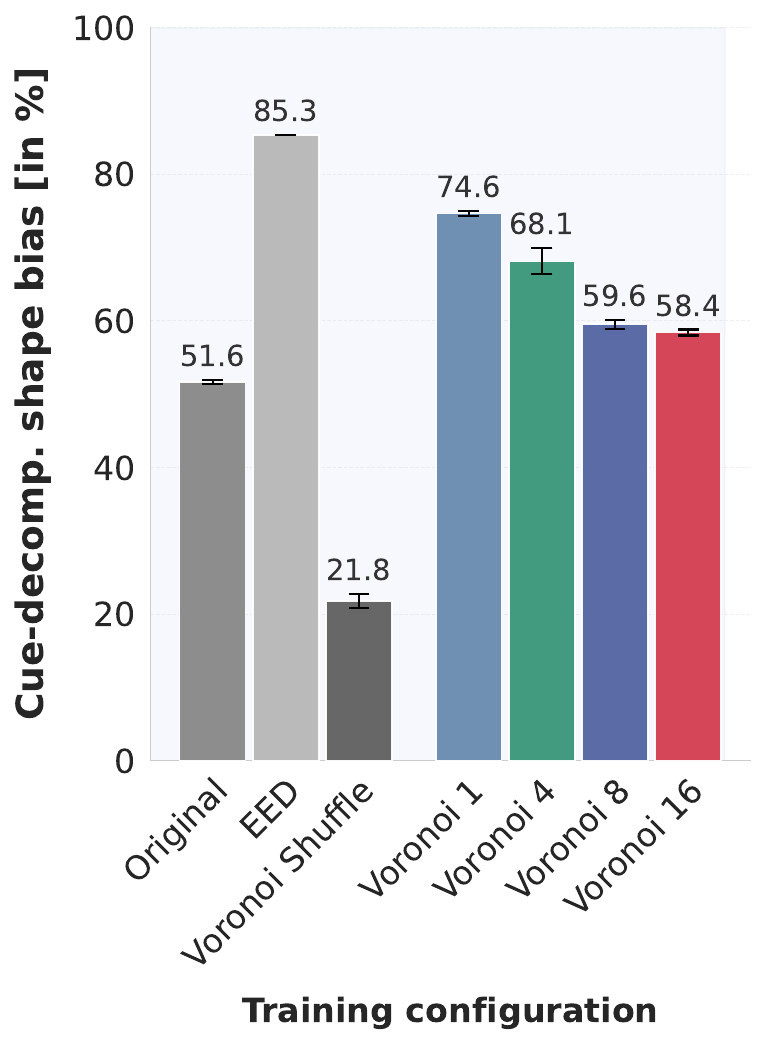}
    \caption{mIoU (left) and CDSB (right) scores of DeepLabV3+ models on the PASCAL Context test subset. Error bars show mean and standard deviation over two training runs per setting.}
    \label{fig:pc_miou_cdsb_barplot} 
    \vspace{4pt}
\end{figure}

\paragraph{Shape bias evaluation.}
To analyze the influence of the stylization on semantic segmentation we evaluate the model performance in terms of mIoU on the original task (non-transferred data) and measure the shape bias. The effectiveness of the offline data augmentation developed is studied by varying key parameters as described in \cref{sec:exp_setup}.
First, the effect of the number of Voronoi cells $n$ is studied, ensuring each cell is stylized ($p = 1$). Afterwards, the stylization proportion is varied on the best-performing network w.r.t.\ the number of Voronoi cells.
We start by investigating mIoU values on original data of differently trained models. The results for Cityscapes are depicted in \Cref{fig:cs_miou_barplot} and in \Cref{fig:pc_miou_cdsb_barplot} (left) for PASCAL Context.

Compared to the model trained on the original data, all networks trained with one of the augmentation methods lose mIoU performance due to the domain gap introduced thereby. The EED baselines suffer the strongest whereas stylization (ours) still shows descent performance, indicating a proper generalization capability to the original domain. For Cityscapes we investigated partly stylization ($p <1$). When only up to 50\% of the cells are stylized (Voronoi 16 25\% and Voronoi 16 50\%) the performance is nearly comparable with the original performance. It follows, that the parameter $p$ manages the trade-off between reducing texture bias and preserving original texture to sustain high mIoU on the original dataset. However, we see a clear difference between network architectures. While the CNN-based architecture on Cityscapes benefits from an increased number of Voronoi cells and therefore more different styles per image, the transformer model shows the best performance when trained on images augmented with a single style per image, (i.e., Voronoi 1). For PASCAL Context we observe that texture is highly discriminatory as a model trained on Voronoi-shuffled data achieves relatively high mIoU performance.
This is mirrored in the stylization where in contrast to Cityscapes the single style per image augmentation achieves higher mIoU performance.

\begin{figure}[tbp]
    \centering
    \includegraphics[width=\columnwidth]{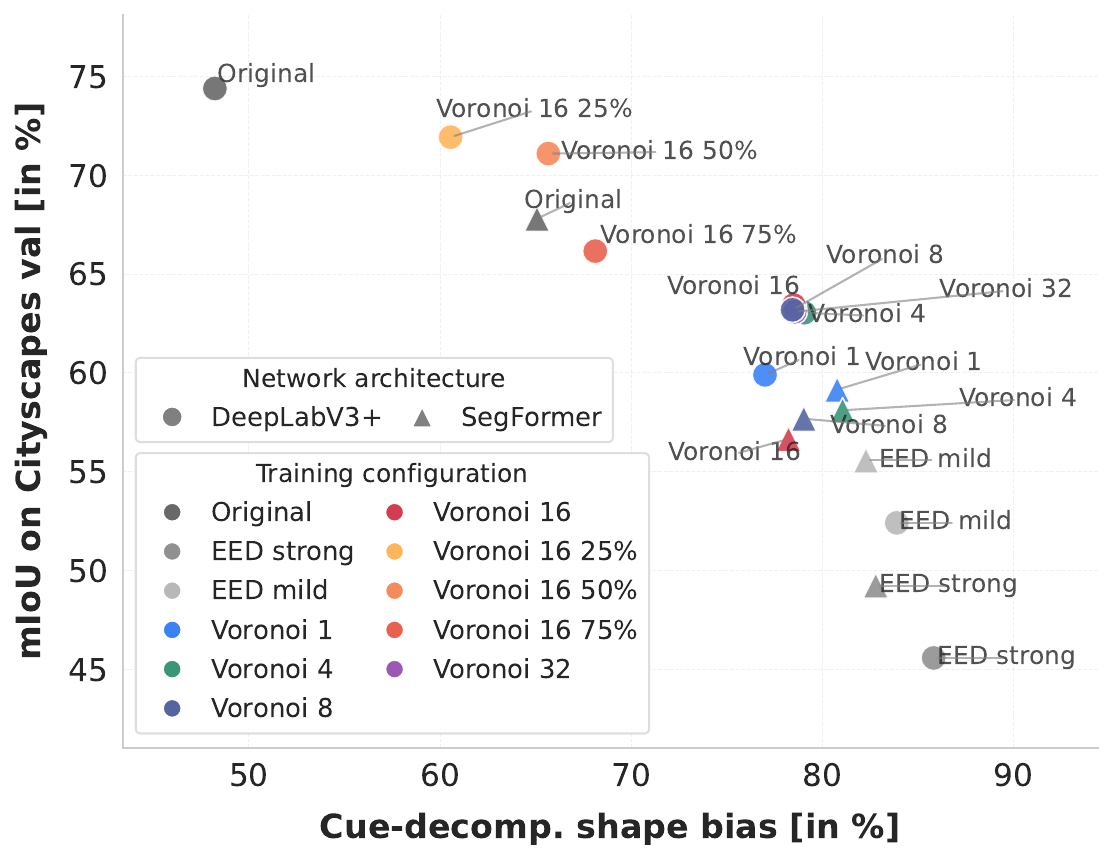}
    \caption[Scatterplot CDSB vs. mIoU on Cityscapes]{Scatterplot of CDSB vs.\ mIoU on Cityscapes validation data for various augmentation settings.}
    \label{fig:scatterplot_cdsb_miou}
    \vspace{4pt}
\end{figure}

We further analyze whether and to which extent the texture bias can be reduced or alternatively the shape bias can be increased by (cell-) stylization. 
The shape bias is measured in terms of CDSB \eqref{eq:cdsb} for each of the previously introduced training configurations and presented in \Cref{fig:cs_cdsb_barplot} for Cityscapes, in the same way as the mIoU comparison before. As expected from the construction of the benchmark (which uses EED for shape-bias assessment), the EED-baseline models achieve the highest CDSB scores for both architectures, as one component of the score is calculated as the mIoU on images augmented with the $\text{EED}_{\text{mild}}$ parameters. For this reason, the $\text{EED}_{\text{mild}}$ models can be viewed as oracles in this experiment. Notably, all networks trained on stylized data exceed the shape bias of the original data baseline network. 
Also, we see clear differences in how both network architectures perform, especially the shape bias of the baseline SegFormer network (original) is much higher than the CDSB of the baseline DeeplabV3+, which is in line with results found in literature \cite{tuli2021convolutionalneuralnetworkstransformers, heinert2025shape}. Training the CNN-based segmentation model with stylized Cityscapes data results in more than $30$ percent points gain in the shape bias measure for all fully stylized approaches, highest for $n=4$. However, the CDSB values are still slightly below those of the SegFormer trained on the same stylized data, only the EED oracles achieve a higher shape bias.

Furthermore, adjusting the proportion $p$ of stylized segments reveals a favorable and well-controllable trade-off between mIoU on original images and shape bias. The interplay between the model performance and CDSB scores for Cityscapes are visualized in \Cref{fig:scatterplot_cdsb_miou}. The models trained on images with $p\in\{0.5,0.75\}$ achieve CDSB scores closer to those of models trained on fully stylized images than to the baseline model trained on original images whereas the mIoU on original images remains closer to that of the baseline model.
Furthermore, the plot shows that the two metrics are negatively correlated for both model architectures.
The results for the CNN indicate that its shape bias benefits from the introduction of multiple Voronoi cells per image. Choosing $n > 4$ does not yield further gains in shape bias, similarly to the mIoU on the original data where higher numbers of Voronoi cells show almost equal results. Further discussion of the CDSB components is provided in the appendix B.
The SegFormer models show a slight decrease in shape bias when increasing the number of Voronoi cells. The strongest results are achieved for stylization with $n=1$ and $n=4$ stylized cells, achieving nearly the same CDSB scores as the EED oracle but with distinctly better mIoU performance on the original task.
For PASCAL Context (see \Cref{fig:pc_miou_cdsb_barplot} (right)) with increasing Voronoi cells the CDSB score decreases. This arises as not only the performance on shape data ($\text{IoU}_\text{Shape}$) increases compared to the original performance but also the gap to texture data reduces with increasing cells (increase of $\text{IoU}_\text{Texture}$). However, all stylizations improve the shape bias compared to original and texture data (Voronoi shuffle).   

\paragraph{Analysis of IoU per class.}
To investigate whether certain classes suffer or benefit more from stylization, we exemplary compare per-class IoUs of baseline models and the models trained on stylized images which achieved the highest $\text{IoU}_\text{Shape}$ in CDSB for Cityscapes. The result tables with the complete class lists are displayed and discussed in more detail in appendix C.
The relative order of class IoUs within model architectures between baseline networks and those trained on stylized data remains almost identical, except for the regime of the less frequent classes. Furthermore, the rare classes in terms of pixel count also show the highest mIoU performance drop under stylization, except for rider. This behavior is observed for both model architectures.
For PASCAL Context, we observe that classes suffer differently from the stylization. While \textit{areoplane}, \textit{boat} and \textit{sheep} lose roughly 20 percent points in IoU, classes like \textit{bicycle} and \textit{diningtable} lose less than 3 percent points. 
It seems that there is no semantic grouping for the classes which suffer less or more by stylization. However, the results for the salient classes generalize across the dataset as observed for \textit{train}, \textit{bus} and \textit{motorcycle}.

\begin{figure*}[t]
    \centering
    \includegraphics[trim=.3cm 0 0.5cm 0,clip,width=0.305\linewidth]{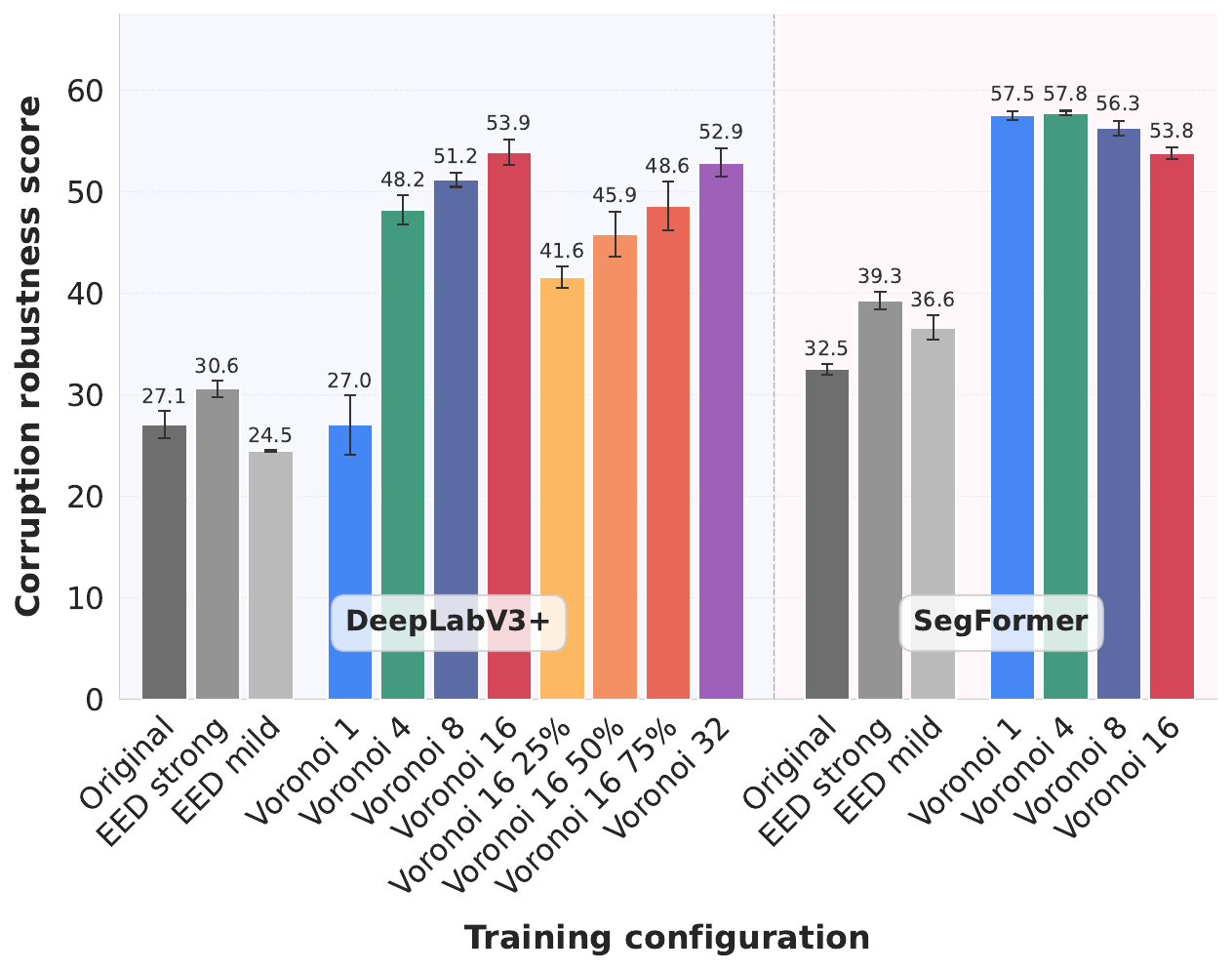}
    \hspace{0.2ex}
    \textcolor{gray}{\rule{0.5pt}{4.5cm}}
    \hspace{0.2ex}
    \includegraphics[trim=.3cm 0 0.5cm 0,clip,width=0.32\linewidth]{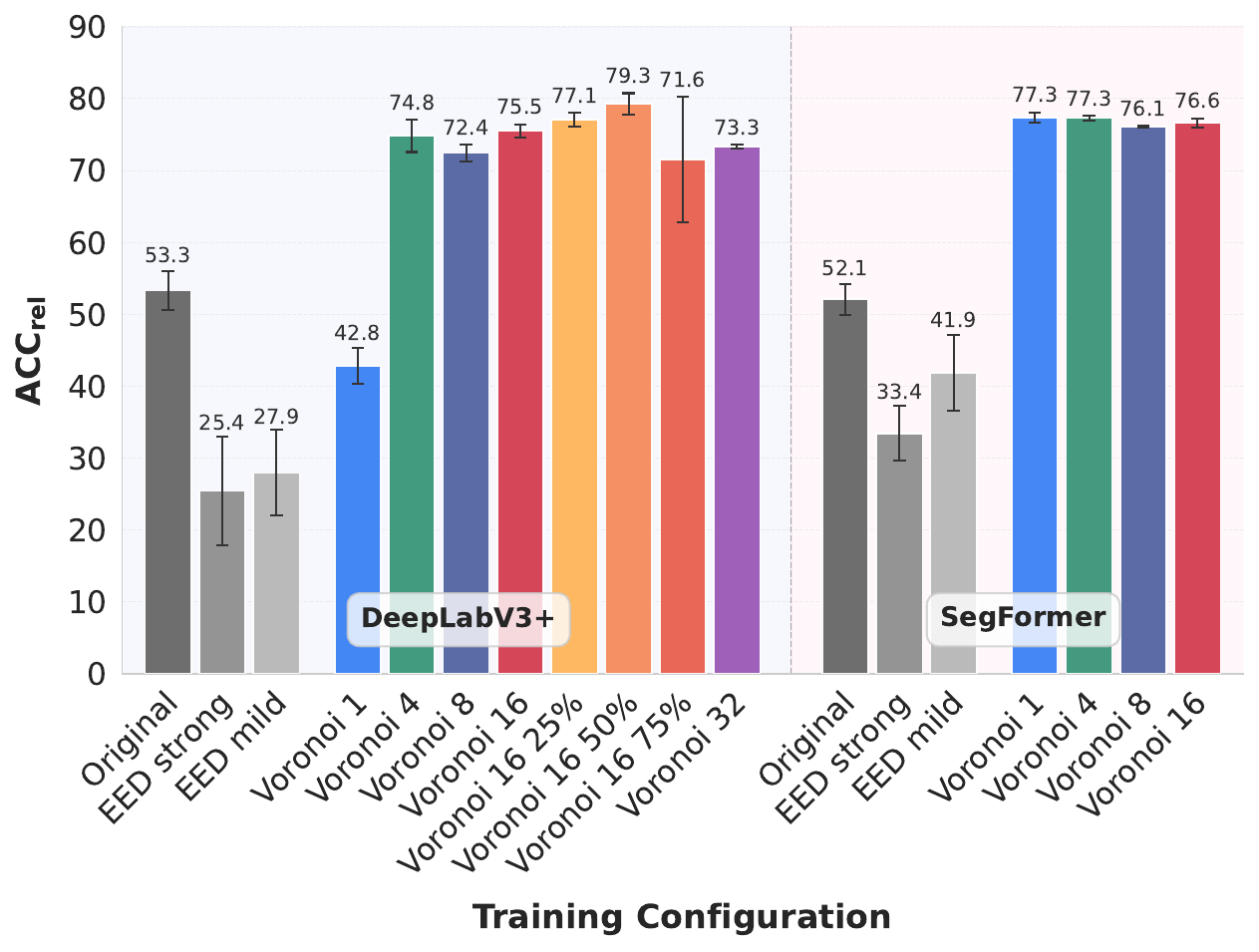} 
    \includegraphics[trim=.3cm 0 0.5cm 0,clip,width=0.32\linewidth]{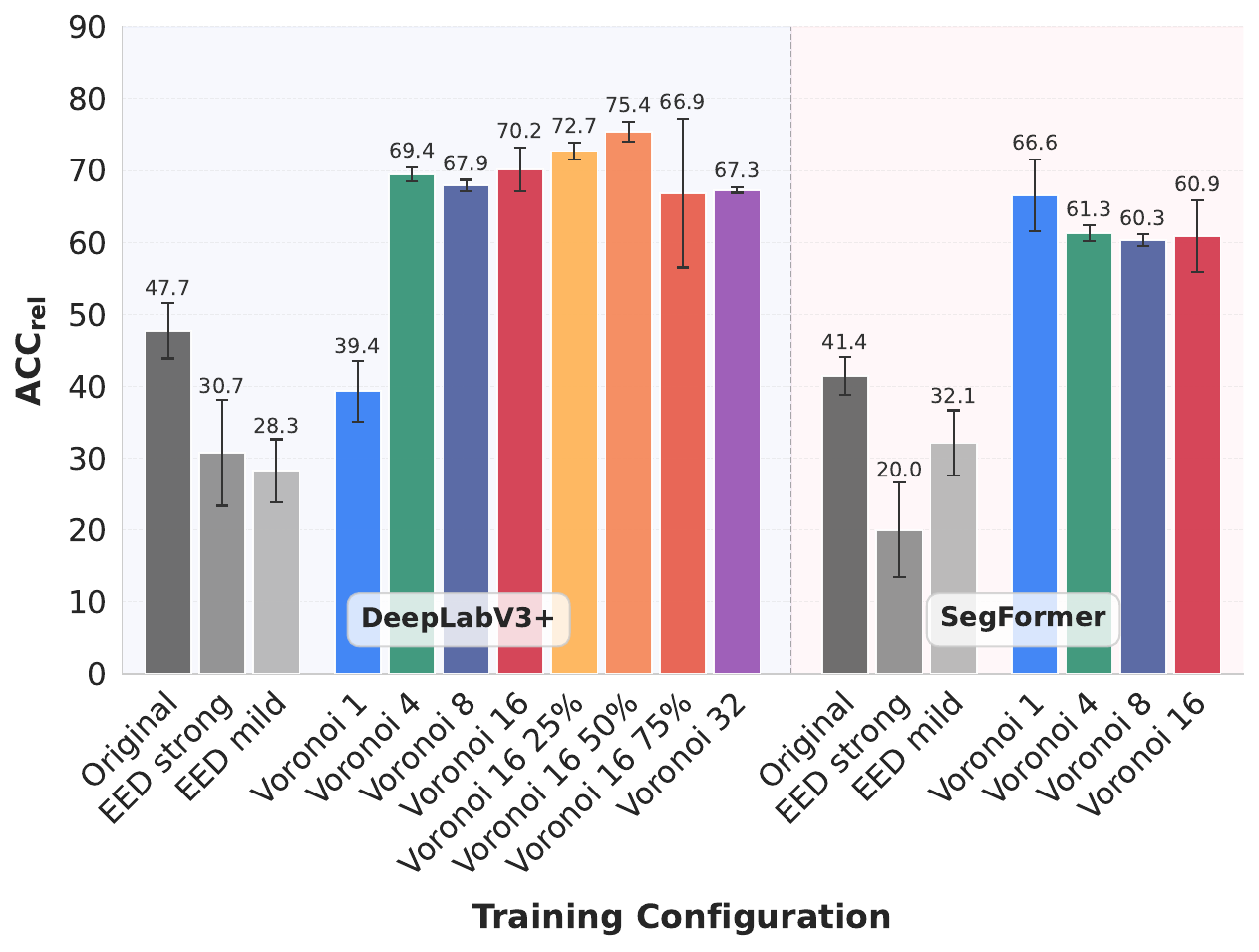}
    \caption{Visualization of image corruption robustness scores with error bars to display standard deviations (left). Robustness results for untargeted (center) and targeted (right) FGSM attacks on Cityscapes images. The error bars show the mean and standard deviation of the mean relative accuracy across attack strength over all three models per training configuration.}
    \label{fig:fgsm_corruption_barplot} 
\end{figure*}

\paragraph{Corruption robustness results.}
One desirable property in segmentation models is robustness to input perturbations. In our case, this means predictions should remain stable under slight image corruptions. Models with a stronger shape bias have been found to be more robust, and human vision is very robust when faced with image corruptions~\cite{Geirhos2018ImageNettrainedCA}. 
We evaluate trained networks on different distorted versions of the Cityscapes validation dataset, i.e., one version for each corruption type and intensity value. The corruption robustness score is calculated as the mean of mIoUs achieved, where the mean is computed over all corruptions and intensities divided. This mean is divided by the mIoU on the original dataset and summarizes the relative robustness to image corruptions as described in \cref{sec:exp_setup}. Additional per-corruption-type analysis and plots can be found in appendix E. Overall, both DNNs profit highly from style transfer augmentation, see \Cref{fig:fgsm_corruption_barplot}. For DeepLabV3+, it increases from $27.08$ (baseline) to $53.91$ ($+99.06$\%) using Voronoi 16, while for SegFormer, it rises from $32.54$ to $57.80$ ($+77.63$\%) with Voronoi 4.

\begin{figure}[tb]
    \centering
    \setlength{\tabcolsep}{1.5pt}
    \renewcommand{\arraystretch}{0.5}
        \begin{tabular}{ccc}
         Input image & DL original baseline & DL Voronoi 16 \\
         \includegraphics[width=0.32\columnwidth]{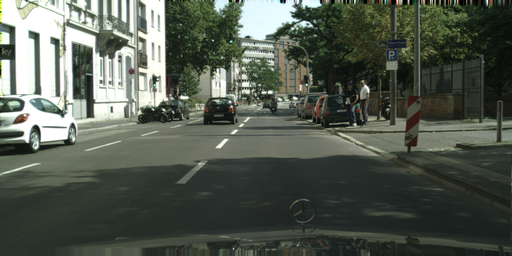} & \includegraphics[width=0.32\columnwidth]{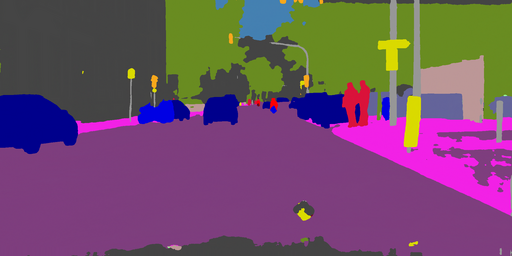} & \includegraphics[width=0.32\columnwidth]{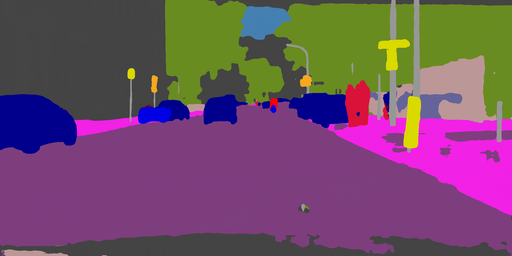}\\
         \includegraphics[width=0.32\columnwidth]{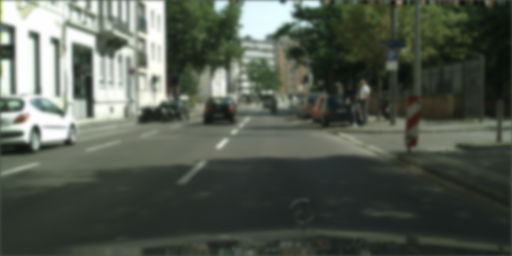} & \includegraphics[width=0.32\columnwidth]{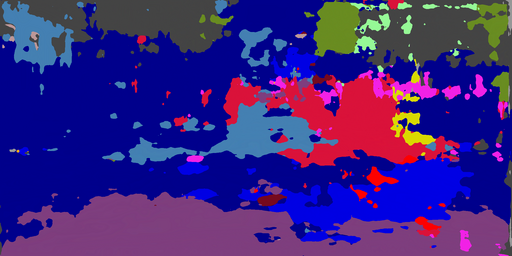} & \includegraphics[width=0.32\columnwidth]{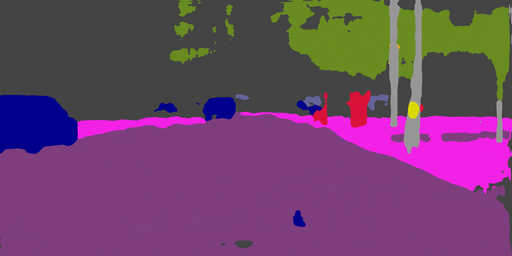} \\
        \includegraphics[width=0.32\columnwidth]{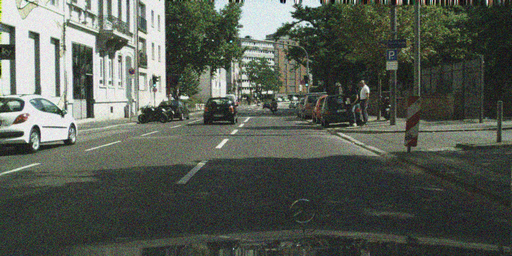} & \includegraphics[width=0.32\columnwidth]{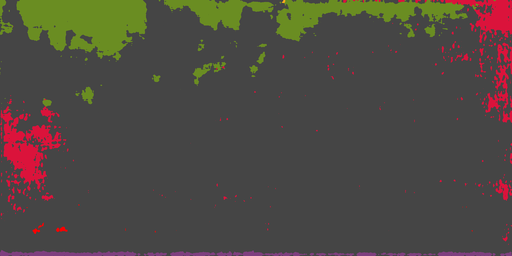} & \includegraphics[width=0.32\columnwidth]{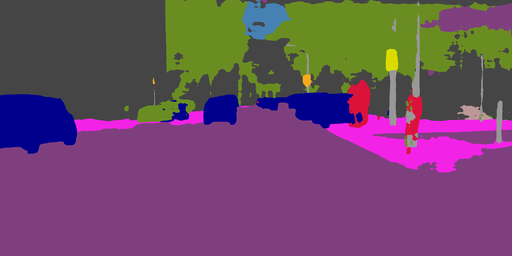} \\
         \includegraphics[width=0.32\columnwidth]{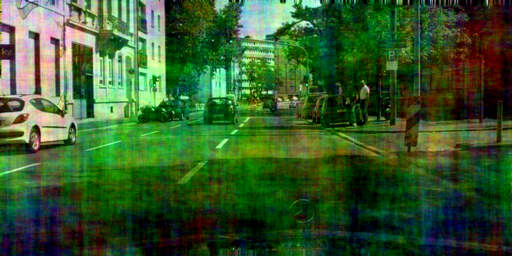} & \includegraphics[width=0.32\columnwidth]{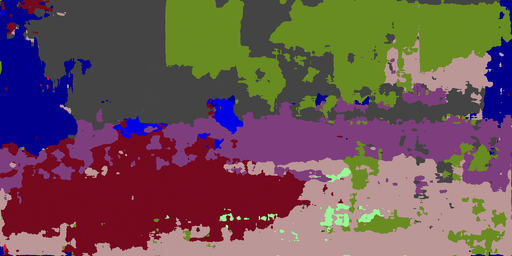} & \includegraphics[width=0.32\columnwidth]{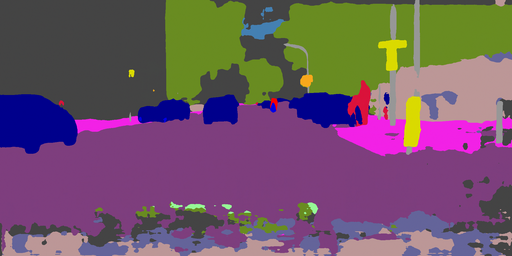}
    \end{tabular}
    \caption[Image Corruptions Visualization]{Illustration of image corruptions and predictions of CNN-based networks trained with and without style-transfer augmentation. Corruptions are low-pass filtering with $\sigma=7.0$, uniform noise with $\eta=0.35$ and phase noise with $w=60$.}
    \vspace{4pt}
\label{fig:corruption_visualization}
\end{figure}

Visual examples of low pass, uniform noise and phase noise corruptions alongside the predictions of CNN-
based networks trained with and without style-transfer augmentation are presented in \Cref{fig:corruption_visualization} and \Cref{fig:fgsm_corruption_barplot} (left) illustrates the robustness scores of the trained networks. SegFormer models that were not trained on EED or style transfer data exhibit higher robustness to image corruptions than DeepLabV3+, consistent with their higher shape bias observed earlier. Training on EED-processed data yields moderate improvements for SegFormer, while DeepLabV3+ shows only minor gains, even with stronger EED settings.
Furthermore, both model architectures react differently to the stylization of input images. The convolutional network performs similarly to the original image baseline when trained on images stylized with only one style. Further increasing the number of Voronoi cells $n$ yields higher gains in corruption robustness up to $n=16$, beyond which robustness does not improve. Partial stylization of training data significantly improves robustness in all variants compared to the baseline model, with the Voronoi 16 75\% model being more robust than the Voronoi 4 model. The previously identified favorable trade-off between mIoU on the original dataset and CDSB can be observed again for robustness to image corruptions. Here, networks with partial stylization are closer to the fully stylized variants than to the baseline in terms of robustness to image corruptions. Notably, the standard deviations of the robustness score are slightly higher in DeepLabV3+ models, which indicates that the robustness of trained convolutional models could be a less stable property than it is in SegFormer models.

For SegFormer models, stylizing the entire image with one or four styles yields the strongest robustness results, which is again in line with our findings in the shape bias benchmark. In contrast, increasing the segment count leads to lower robustness scores.
Overall, the transformer architecture itself comes with a slightly higher robustness when faced with corruptions, but the convolutional architecture shows drastic increases with style-transfer-based training data augmentation, which significantly narrows the gap.

\paragraph{Robustness to adversarial attacks.}
A natural question that arises is whether our method also improves robustness against adversarial attacks, i.e., small perturbations (imperceptible to humans) added into the input images that lead a DNN to make incorrect predictions during testing time \cite{Maag_2024_WACV}. They pose a significant challenge and emphasize the importance of robustness to such adversarial examples \cite{Maag2023,Xu2021}.
As attack, we consider the widely used single-step \emph{fast gradient sign method} (FGSM, \cite{Goodfellow2015}) applied pixel-wise in two versions, untargeted and targeted (the “least likely” class is chosen as target) \cite{Arnab2020}, using a strength of the perturbation of $\{ 2,4,8,16,32 \}$. 
As the performance values obtained for Cityscapes vary for training on different data, we introduce a relative accuracy metric to ensure a fair comparison of the approaches. This metric is given by 
\begin{equation}\label{eq:apsr}
    \text{ACC}_{\mathit{rel}} = \frac{\text{ACC}_{\mathit{AA}}}{\text{ACC}_{\mathit{CS}}}
\end{equation}
where $\text{ACC}_{\mathit{CS}}$ denotes the accuracy of the model evaluated on Cityscapes and $\text{ACC}_{\mathit{AA}}$ after performing the adversarial attack. 

In \Cref{fig:fgsm_corruption_barplot}, the evaluation results for both types of FGSM on the differently trained networks are given. Additionally, detailed per-attack-strength visualizations are provided in Appendix F.
For both attacks and models, the EED baseline models show the lowest robustness against the attacks tested, followed by the original data baseline models, which consistently (except for Voronoi 1 DeepLabV3+) perform worse than the stylized trained models. 
For the DeepLabV3+, the model trained with 16 stylized segments achieves the strongest results.  
Interestingly, the models trained on partially stylized images with $p\in\{0.25,0.5\}$ present the highest levels of robustness to adversarial attacks observed in the experiments, with higher values of relative accuracy compared to fully stylized approaches. 
The SegFormer models also benefit from stylized training images, but the exact number of Voronoi cells does not show a strong impact on the robustness to adversarial attacks. 
In general, the models show higher accuracy in untargeted attacks, especially the SegFormer models appear to be more prone to targeted attacks. 
In both plots, the standard deviations are higher than before in many cases, indicating that the robustness to adversarial attacks may be more difficult to control reliably via stylizing training images than the shape bias and robustness to image corruptions measured previously. In any case, the results indicate that the introduction of style transfer into the training images has a strong positive effect on robustness to adversarial attacks.

\section{Conclusion}\label{sec:outlook}

In this paper, we provided the first study on the influence of style transfer on the reduction of texture bias and robustness w.r.t.\ image corruptions in semantic segmentation. This study gives insights into learned cue biases of semantic segmentation models and how to enhance the reliance of semantic segmentation models on shape-based features. In an offline augmentation procedure, we transferred the style of arbitrary paintings to semantic segmentation datasets. We investigated the stylization of entire images as well as applying multiple different styles per image by stylizing random Voronoi cells.
Our study reveals that both architectures -- convolution-based and transformer-based -- benefit from the offline style augmentation w.r.t.\ shape bias and robustness.
In detail, cell-wise style transfer boosts shape bias and robustness in CNNs. However, stylization leads to reduced mIoU performance. Partial stylization ($p < 1$) allows for balancing the original segmentation performance and shape bias as well as the robustness against image corruptions and adversarial attacks.
The SegFormer, being less texture-biased from the inception of its design, 
exhibits diminished bias shifts. However,
SegFormer benefits from style transfer, best with single-style global stylization.
In addition, our study reveals that shape bias correlates with corruption robustness for stylized models. The increased shape bias particularly improves the robustness with respect to untargeted as well as targeted FGSM adversarial attacks with diverse attack strengths.
This simple, yet effective, offline augmentation allows to steer the shape bias and improve semantic segmentation robustness with respect to image corruptions as well as adversarial attacks.

\section*{Acknowledgment}
A.M.\, E.H.\ and M.R.\ acknowledge support by the German Federal Ministry of Education and Research (BMBF) within the junior research group project ``UnrEAL'' (grant no.\ 01IS22069).

\includegraphics[width=4cm]{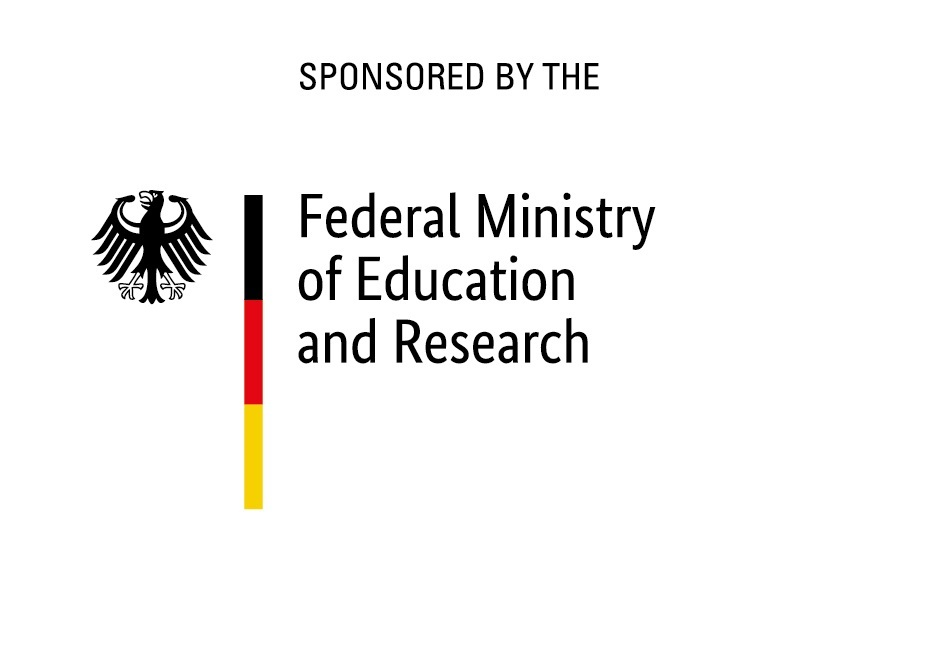}

{
    \small
    \bibliographystyle{IEEEtran}
    \bibliography{mybibfile}
}

%%%%%%%%%%%%%%%%%%%%%%%%%%%%%%%%%%%%%%%%%%%%%%%%%%%%%%%%%%%%%%%%% 
%%%%%% Appendix.
\newpage
\appendices
In this appendix, we provide supplementary material referenced throughout the main paper. It includes technical details of the procedures, additional visualizations and detailed tables to provide further insight into the research and analysis presented.

\section{Tested offline augmentations}

\begin{figure}[htbp]
\centering
\setlength{\tabcolsep}{1pt}
\begin{tabular}{cc}

    \begin{minipage}[t]{0.24\textwidth}
        \includegraphics[width=\textwidth]{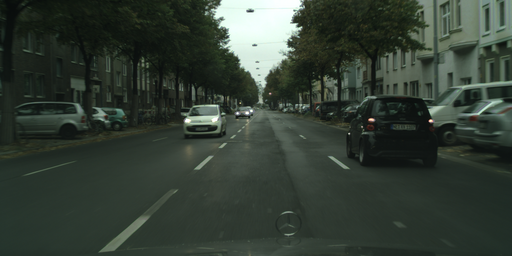}
        \centerline{\scriptsize Original}
    \end{minipage} &
    \begin{minipage}[t]{0.24\textwidth}
        \includegraphics[width=\textwidth]{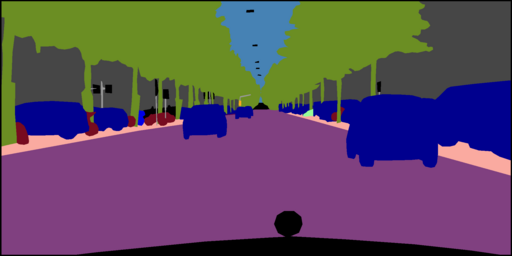}
        \centerline{\scriptsize Ground Truth}
    \end{minipage} \\
    \vspace{1pt}

    \begin{minipage}[t]{0.24\textwidth}
        \includegraphics[width=\textwidth]{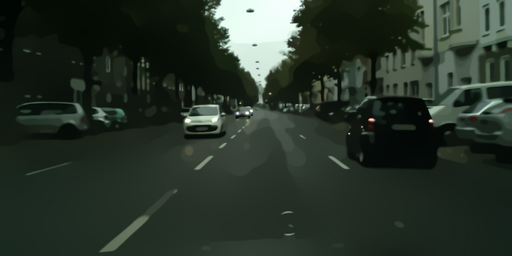}
        \centerline{\scriptsize $\text{EED}_{\text{mild}}$}
    \end{minipage} &
    \begin{minipage}[t]{0.24\textwidth}
        \includegraphics[width=\textwidth]{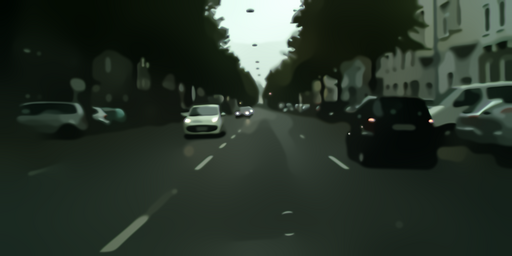}
        \centerline{\scriptsize $\text{EED}_{\text{strong}}$}
    \end{minipage} \\
    \vspace{1pt}

    \begin{minipage}[t]{0.24\textwidth}
        \includegraphics[width=\textwidth]{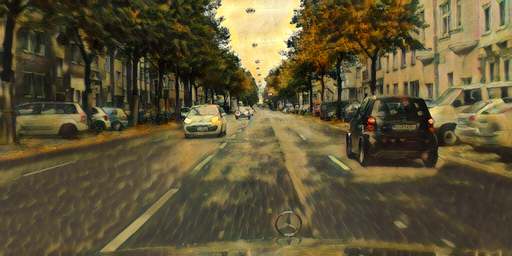}
        \centerline{\scriptsize Stylized ($n=1$)}
    \end{minipage} &
    \begin{minipage}[t]{0.24\textwidth}
        \includegraphics[width=\textwidth]{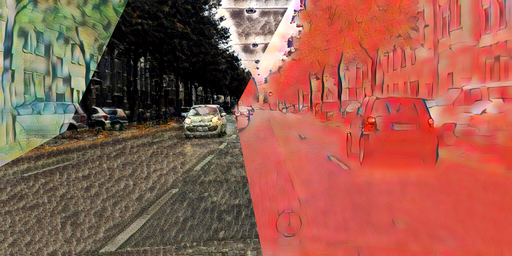}
        \centerline{\scriptsize Voronoi 4}
    \end{minipage} \\
    \vspace{1pt}

    \begin{minipage}[t]{0.24\textwidth}
        \includegraphics[width=\textwidth]{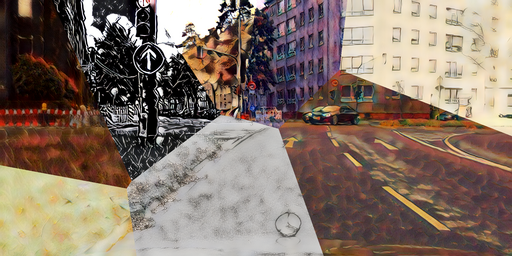}
        \centerline{\scriptsize Voronoi 8}
    \end{minipage} &
    \begin{minipage}[t]{0.24\textwidth}
        \includegraphics[width=\textwidth]{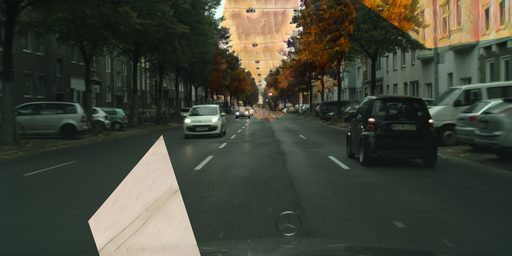}
        \centerline{\scriptsize Voronoi 16 (25\%)}
    \end{minipage} \\
    \vspace{1pt}

    \begin{minipage}[t]{0.24\textwidth}
        \includegraphics[width=\textwidth]{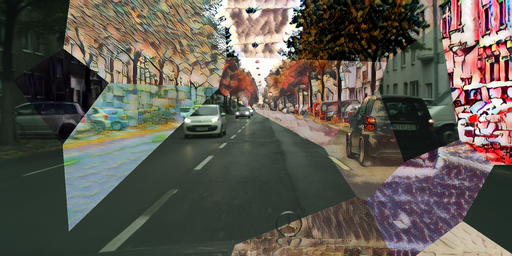}
        \centerline{\scriptsize Voronoi 16 (50\%)}
    \end{minipage} &
    \begin{minipage}[t]{0.24\textwidth}
        \includegraphics[width=\textwidth]{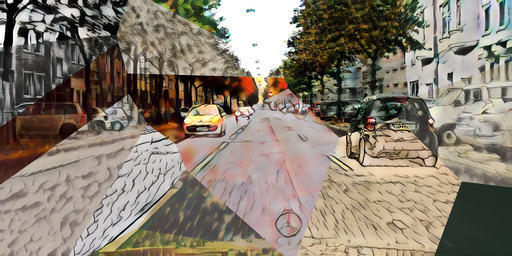}
        \centerline{\scriptsize Voronoi 16 (75\%)}
    \end{minipage} \\
    \vspace{1pt}

    \begin{minipage}[t]{0.24\textwidth}
        \includegraphics[width=\textwidth]{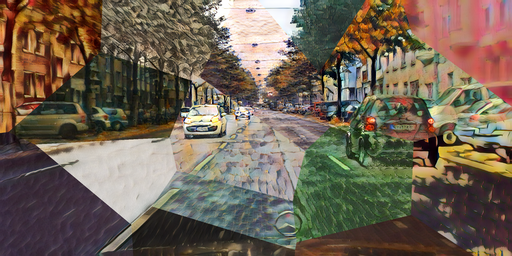}
        \centerline{\scriptsize Voronoi 16}
    \end{minipage} &
    \begin{minipage}[t]{0.24\textwidth}
        \includegraphics[width=\textwidth]{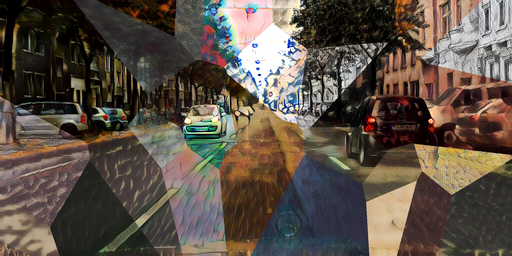}
        \centerline{\scriptsize Voronoi 32}
    \end{minipage}
\end{tabular}
\caption[Visualization of augmented training images]{Visualization of all analyzed training datasets, original image and ground truth mask sourced from Cityscapes~\cite{cordts2016cityscapesdatasetsemanticurban}, EED images as proposed by \cite{heinert2024reducingtexturebiasdeep}.}
\label{fig:train_augmentations}
\end{figure}

\Cref{fig:train_augmentations} displays all training data augmentations used in our Cityscapes experiments. We investigate the parameters $n$ and $p$ in a hyperparameter study. Note that the number of stylized patches may not exactly correspond to the stylization proportion $p$ due to the implementation of partial stylization as a Bernoulli distribution over the Voronoi patches. 
We measured the runtime of our patch-wise style transfer with varying numbers of Voronoi cells on an NVIDIA Quadro P6000 GPU and present the results in \cref{tab:stylization_runtime}. While regular AdaIN is fast, the added Voronoi overhead increases with cell count. This makes the method impractical for online use but suitable for offline precomputation.

\begin{table}[h]
\centering
\caption{Stylization runtime (in seconds) per image with varying numbers of Voronoi cells, measured on an NVIDIA Quadro P6000. 10 runs per Stylization method.}
\label{tab:stylization_runtime}
\begin{tabular}{lccc}
\toprule
\textbf{Style Transfer Method} & \textbf{Voronoi Cells} & \textbf{Mean (s)} & \textbf{Std (s)} \\
\midrule
AdaIN (no Voronoi)     & 1   & 0.163 & 0.037 \\
Voronoi stylization            & 4   & 1.095 & 0.027 \\
                               & 8   & 2.070 & 0.014 \\
                               & 16  & 4.088 & 0.018 \\
                               & 32  & 8.168 & 0.019 \\
\bottomrule
\end{tabular}
\end{table}

\section{CDSB component dissection}
\begin{table*}[htbp]
\caption[Dissection of Shape Bias Components]{Detailed shape bias and performance metrics across Cityscapes evaluation datasets. The scores represent mIoU values on the evaluation data which result from averaging over training runs, 
complemented with the respective standard deviations over those runs}.
\label{tab:shape_bias_detailed}
\centering
\resizebox{0.8\textwidth}{!}{
\begin{tabular}{lccccc}
\toprule
\textbf{Training Configuration} & \textbf{Architecture} & \textbf{CDSB Score} & \textbf{Cityscapes val} & \textbf{Voronoi Shuffle 128}& $\textbf{EED}_\textbf{mild}$ \\
\midrule
Original & DeepLabV3+ & 0.48 $\pm$ 0.03 & 74.40 $\pm$ 0.56 & 29.83 $\pm$ 1.09 & 19.90 $\pm$ 2.48 \\
EED strong & DeepLabV3+ & 0.86 $\pm$ 0.01 & 45.57 $\pm$ 1.28 & 11.66 $\pm$ 0.64 & 50.34 $\pm$ 0.24 \\
EED mild (oracle) & DeepLabV3+ & 0.84 $\pm$ 0.00 & 52.41 $\pm$ 1.11 & 15.13 $\pm$ 0.21 & 56.23 $\pm$ 0.59 \\
Voronoi 1 & DeepLabV3+ & 0.77 $\pm$ 0.01 & 59.90 $\pm$ 2.07 & 13.75 $\pm$ 0.46 & 32.90 $\pm$ 2.24 \\
Voronoi 4 & DeepLabV3+ & 0.79 $\pm$ 0.00 & 63.05 $\pm$ 0.43 & 15.19 $\pm$ 0.14 & 40.93 $\pm$ 1.01 \\
Voronoi 8 & DeepLabV3+ & 0.78 $\pm$ 0.00 & 63.19 $\pm$ 1.14 & 15.44 $\pm$ 0.35 & 40.08 $\pm$ 0.85 \\
Voronoi 16 & DeepLabV3+ & 0.79 $\pm$ 0.01 & 63.44 $\pm$ 0.99 & 15.90 $\pm$ 0.80 & 41.42 $\pm$ 0.97 \\
Voronoi 16 25\% & DeepLabV3+ & 0.61 $\pm$ 0.02 & 71.93 $\pm$ 3.16 & 23.31 $\pm$ 1.02 & 25.52 $\pm$ 0.84 \\
Voronoi 16 50\% & DeepLabV3+ & 0.66 $\pm$ 0.02 & 71.10 $\pm$ 0.55 & 20.80 $\pm$ 0.12 & 28.47 $\pm$ 2.35 \\
Voronoi 16 75\% & DeepLabV3+ & 0.68 $\pm$ 0.02 & 66.16 $\pm$ 3.14 & 19.64 $\pm$ 0.75 & 30.04 $\pm$ 3.01 \\
Voronoi 32 & DeepLabV3+ & 0.79 $\pm$ 0.01 & 63.09 $\pm$ 1.56 & 15.12 $\pm$ 0.52 & 39.52 $\pm$ 1.23 \\
Original & SegFormer & 0.65 $\pm$ 0.01 & 67.80 $\pm$ 0.32 & 24.49 $\pm$ 1.10 & 32.54 $\pm$ 0.55 \\
EED strong & SegFormer & 0.83 $\pm$ 0.01 & 49.23 $\pm$ 0.56 & 14.74 $\pm$ 0.69 & 50.59 $\pm$ 0.35 \\
EED mild (oracle) & SegFormer & 0.82 $\pm$ 0.00 & 55.57 $\pm$ 0.04 & 16.37 $\pm$ 0.33 & 54.22 $\pm$ 0.58 \\
Voronoi 1 & SegFormer & 0.81 $\pm$ 0.01 & 59.15 $\pm$ 0.70 & 14.50 $\pm$ 0.32 & 43.48 $\pm$ 1.53 \\
Voronoi 4 & SegFormer & 0.81 $\pm$ 0.00 & 58.11 $\pm$ 0.41 & 14.16 $\pm$ 0.22 & 43.25 $\pm$ 1.00 \\
Voronoi 8 & SegFormer & 0.79 $\pm$ 0.00 & 57.68 $\pm$ 0.30 & 15.05 $\pm$ 0.19 & 40.47 $\pm$ 0.54 \\
Voronoi 16 & SegFormer & 0.78 $\pm$ 0.00 & 56.65 $\pm$ 0.42 & 15.21 $\pm$ 0.25 & 38.99 $\pm$ 0.46 \\
\bottomrule
\end{tabular}}
\end{table*}

\Cref{tab:shape_bias_detailed} displays the results for each individual evaluation contributing to the CDSB alongside the achieved mIoU on the Cityscapes validation set. The observed numbers largely confirm the concluded results from the key metric evaluations and highlight that the CDSB measure permits the comparison of networks across performance levels. For example, the $\text{EED}_\text{strong}$ networks show a worse performance on Voronoi-shuffled and on $\text{EED}_\text{mild}$ data than the $\text{EED}_\text{mild}$-trained networks, but obtain equal or higher shape bias scores nonetheless. Standard deviations are observed to be rather low for most training configurations, which indicates a relatively stable reaction to the applied training data augmentations.
As expected, baseline models trained on the original dataset achieve the highest mIoUs on the Voronoi-shuffled dataset, which measures prediction quality achieved only from texture information.
In the case of DeepLabV3+ models, it appears that any stylization of training images improves the extrapolation to EED data with the highest gains up to 16 stylized regions. The Voronoi 4 variant appears to exhibit the strongest shape bias but in terms of absolute performance on EED-preprocessed images it is exceeded by the models trained on stylized images with 16 segments. The Voronoi 16 training configuration achieves the highest combined evaluation scores by a small margin over $n\in\{4,8,32\}$ stylized segments, with the Voronoi 4 variant presenting the highest shape bias. Analyzing the results for networks trained on partially stylized images, an increase in mIoU on EED images can be observed, accompanied by a simultaneous decrease in mIoU on Voronoi-shuffled images. This indicates that the measured shape bias and texture bias are complementary properties, as suggested in \cite{heinert2025shape}. In summary, the convolutional network strongly benefits from segment-wise style transferred training images with 16 Voronoi cells showing the best results on Cityscapes. When the images are only partially stylized, the best trade-off between performance and increased shape bias is achieved by stylizing half of the image in the performed experiments. 

SegFormer models behave as expected from CDSB and original data performance evaluations, with Voronoi 1 in the lead and lowering scores for more segments. The Voronoi 4 variant achieves the highest shape bias score, but lower absolute mIoU on the EED data. 

When comparing between model architectures, it becomes apparent that although both architectures become more shape-reliant through the applied data augmentations, the effects on convolutional networks are a lot stronger, with up to an 80\% increase in the CDSB score. This could be caused by the natively higher shape bias of the SegFormer model. SegFormer models show better extrapolation to EED data in essentially all variants with DeepLabV3+ models gaining mIoU values to almost competitive levels with more Voronoi segments. The performance on Voronoi-shuffled data is similar across model architectures for models trained on stylized images.

\begin{table}[htbp]
\centering
\caption[Comparison of IoU per Class]{Comparison of mIoU and class-wise IoUs achieved by stylized models and baseline models trained on original data. All models are evaluated on the Cityscapes validation dataset. Classes are sorted by frequency in descending order.
}
\label{tab:class_ious}
\resizebox{.48\textwidth}{!}{
\begin{tabular}{llllll}
\toprule
{} & \multicolumn{2}{c}{DeepLabV3+} & \multicolumn{2}{c}{SegFormer} \\
\midrule
Train Conf. &      Original & Voronoi 16 &      Original & Voronoi 1 \\
\midrule
mIoU          & 74.40 $\pm$ 0.56 &   63.44 $\pm$ 0.99 &   67.80 $\pm$ 0.32 &   59.15 $\pm$ 0.70 \\
\hdashline
road          & 97.96 $\pm$ 0.21 &   96.82 $\pm$ 0.15 &  97.44 $\pm$ 0.04 &  96.66 $\pm$ 0.04 \\
building  &  91.96 $\pm$ 0.08 &   88.38 $\pm$ 0.21 &  89.97 $\pm$ 0.03 &  87.21 $\pm$ 0.18 \\
vegetation &   91.80 $\pm$ 0.23 &   89.06 $\pm$ 0.24 &  91.25 $\pm$ 0.04 &  88.67 $\pm$ 0.09 \\
car        &  94.13 $\pm$ 0.48 &     92.00 $\pm$ 0.60 &  91.93 $\pm$ 0.25 &   89.55 $\pm$ 0.10 \\
sidewalk   &  84.16 $\pm$ 1.06 &   76.46 $\pm$ 0.89 &   80.01 $\pm$ 0.15 &  74.62 $\pm$ 0.08 \\
sky        &  94.31 $\pm$ 0.21 &   88.82 $\pm$ 2.65 &  94.41 $\pm$ 0.04 &  92.29 $\pm$ 0.12 \\
person    &  81.66 $\pm$ 0.18 &   73.18 $\pm$ 0.35 &  71.63 $\pm$ 0.14 &  63.82 $\pm$ 1.15 \\
pole      &   66.52 $\pm$ 0.20 &   54.99 $\pm$ 0.67 &   52.13 $\pm$ 0.03 &  43.27 $\pm$ 0.34 \\
terrain   &   57.50 $\pm$ 0.38 &   45.02 $\pm$ 2.73 &  61.20 $\pm$ 0.31 &  53.46 $\pm$ 1.57 \\
fence     &  56.97 $\pm$ 1.03 &   47.67 $\pm$ 1.97 &  46.32 $\pm$ 0.65 &  42.51 $\pm$ 1.82 \\
wall      &   44.30 $\pm$ 4.17 &   35.26 $\pm$ 2.33 &  55.15 $\pm$ 1.92 &  41.26 $\pm$ 2.64 \\
traffic sign &  78.49 $\pm$ 0.23 &   64.97 $\pm$ 0.59 &   64.70 $\pm$ 0.32 &  55.29 $\pm$ 0.32 \\
bicycle   &   76.30 $\pm$ 0.27 &   68.02 $\pm$ 0.68 &  67.08 $\pm$ 0.64 &  61.47 $\pm$ 0.44 \\
truck       &  58.95 $\pm$ 8.09 &   54.13 $\pm$ 5.24 &  55.87 $\pm$ 0.94 &  38.18 $\pm$ 8.18 \\
bus       &  82.75 $\pm$ 0.96 &   56.59 $\pm$ 6.49 &   66.67 $\pm$ 1.65 &  54.01 $\pm$ 3.13 \\
traffic light &  70.63 $\pm$ 0.35 &   52.57 $\pm$ 1.91 &  54.45 $\pm$ 0.15 &  40.53 $\pm$ 0.37 \\
train    &   64.70 $\pm$ 6.19 &  35.02 $\pm$ 10.52 &  55.11 $\pm$ 3.23 &  39.86 $\pm$ 6.19 \\
rider     &  60.74 $\pm$ 1.03 &   52.52 $\pm$ 0.42 &  44.96 $\pm$ 0.55 &  33.72 $\pm$ 1.34 \\
motorcycle &  59.69 $\pm$ 0.69 &   33.85 $\pm$ 1.47 &  47.99 $\pm$ 0.99 &  27.49 $\pm$ 3.62 \\

\bottomrule
\end{tabular}}
\end{table}

\begin{table*}
\caption{Class-wise performances of DeeplabV3+ models trained with different augmentation methods and evaluated on the original PASCAL Context dataset. Classes are sorted alphabetically. `FG' groups all foreground classes like car and cat whereas in `BG' we group the remaining classes describing background classes like sky and grass.}
\label{tab:class_miou_pc}
\centering
\begin{tabular}{lllllll}
\hline
 class       & baseline         & EED              & Voronoi 1         & Voronoi 4         & Voronoi 8         & Voronoi 16        \\
\hline
 mIoU        & $52.34 \pm 0.44$ & $25.57 \pm 0.21$ & $39.93 \pm 0.24$ & $36.98 \pm 0.53$ & $38.52 \pm 0.07$ & $36.14 \pm 0.71$  \\
 FG mIoU     & $53.72 \pm 1.11$ & $24.75 \pm 0.74$ & $41.49 \pm 1.39$ & $38.42 \pm 1.22$ & $40.48 \pm 1.04$ & $37.57 \pm 0.61$  \\
 BG mIoU     & $49.92 \pm 0.86$ & $27.02 \pm 0.99$ & $37.21 \pm 1.34$ & $34.48 \pm 1.08$ & $35.08 \pm 1.42$ & $33.63 \pm 0.88$  \\
 \hdashline
 aeroplane   & $69.60 \pm 0.63$ & $38.84 \pm 0.16$ & $49.33 \pm 1.87$ & $47.67 \pm 0.96$ & $51.57 \pm 1.39$ & $46.95 \pm 0.74$ \\
 bicycle     & $55.22 \pm 0.31$ & $31.45 \pm 0.84$ & $53.29 \pm 1.82$ & $44.76 \pm 0.41$ & $50.34 \pm 1.99$ & $49.06 \pm 0.77$ \\
 bird        & $45.71 \pm 1.46$ & $14.03 \pm 0.99$ & $28.00 \pm 1.45$ & $29.66 \pm 1.00$ & $25.11 \pm 3.35$ & $24.77 \pm 0.01$ \\
 boat        & $47.34 \pm 0.15$ & $19.18 \pm 1.40$ & $25.51 \pm 1.78$ & $25.41 \pm 0.51$ & $32.02 \pm 0.49$ & $28.99 \pm 0.08$ \\
 bottle      & $33.64 \pm 0.67$ & $15.21 \pm 0.53$ & $27.17 \pm 1.15$ & $20.59 \pm 5.26$ & $22.00 \pm 1.45$ & $23.66 \pm 1.09$ \\
 building    & $42.28 \pm 0.40$ & $23.96 \pm 1.31$ & $32.98 \pm 0.40$ & $30.55 \pm 0.61$ & $32.41 \pm 0.68$ & $29.26 \pm 0.08$ \\
 bus         & $83.00 \pm 1.28$ & $60.67 \pm 1.01$ & $64.84 \pm 2.65$ & $68.09 \pm 1.18$ & $62.65 \pm 0.20$ & $60.84 \pm 0.58$ \\
 car         & $63.88 \pm 1.58$ & $45.09 \pm 0.57$ & $47.57 \pm 1.64$ & $51.64 \pm 2.16$ & $52.02 \pm 1.31$ & $49.62 \pm 0.53$ \\
 cat         & $61.56 \pm 0.47$ & $10.71 \pm 1.52$ & $50.59 \pm 2.76$ & $45.34 \pm 1.38$ & $50.10 \pm 0.59$ & $48.70 \pm 0.56$ \\
 ceiling     & $29.78 \pm 0.89$ & $16.76 \pm 0.48$ & $15.27 \pm 1.75$ & $17.19 \pm 0.43$ & $13.59 \pm 3.58$ & $10.36 \pm 0.88$ \\
 chair       & $22.77 \pm 1.30$ & $11.60 \pm 1.11$ & $17.34 \pm 0.60$ & $14.16 \pm 0.72$ & $15.62 \pm 0.45$ & $15.12 \pm 0.07$ \\
 cow         & $45.70 \pm 2.26$ & $10.12 \pm 0.74$ & $29.82 \pm 2.09$ & $26.48 \pm 1.93$ & $31.75 \pm 1.75$ & $23.00 \pm 0.41$ \\
 diningtabel & $32.00 \pm 0.60$ & $10.10 \pm 0.94$ & $29.09 \pm 1.03$ & $29.84 \pm 1.80$ & $28.75 \pm 0.44$ & $28.86 \pm 0.49$ \\
 dog         & $57.70 \pm 0.39$ & $10.96 \pm 0.28$ & $45.05 \pm 0.19$ & $42.32 \pm 1.11$ & $43.29 \pm 1.59$ & $40.85 \pm 1.05$ \\
 floor       & $46.94 \pm 0.47$ & $10.62 \pm 0.36$ & $32.92 \pm 1.78$ & $32.41 \pm 0.65$ & $34.98 \pm 0.34$ & $30.05 \pm 2.17$ \\
 grass       & $68.06 \pm 0.68$ & $36.16 \pm 0.72$ & $54.27 \pm 1.03$ & $50.14 \pm 0.93$ & $50.47 \pm 1.32$ & $52.64 \pm 0.98$ \\
 ground      & $41.63 \pm 0.79$ & $25.39 \pm 0.59$ & $31.76 \pm 0.15$ & $28.07 \pm 0.23$ & $30.70 \pm 1.25$ & $26.18 \pm 1.56$ \\
 horse       & $55.20 \pm 1.84$ & $9.97 \pm 1.02$  & $44.88 \pm 1.41$ & $40.59 \pm 0.28$ & $44.59 \pm 2.41$ & $38.09 \pm 1.14$ \\
 keyboard    & $33.39 \pm 3.13$ & $15.84 \pm 0.39$ & $17.88 \pm 0.41$ & $15.66 \pm 0.68$ & $26.73 \pm 0.73$ & $9.53 \pm 0.29$  \\
 motorbike   & $61.61 \pm 1.45$ & $34.37 \pm 0.56$ & $52.44 \pm 1.32$ & $45.00 \pm 2.30$ & $49.63 \pm 0.12$ & $47.16 \pm 0.17$ \\
 mountain    & $32.12 \pm 1.73$ & $9.14 \pm 2.24$  & $23.52 \pm 1.12$ & $21.84 \pm 2.85$ & $19.39 \pm 1.90$ & $20.75 \pm 0.23$ \\
 person      & $72.40 \pm 0.23$ & $45.48 \pm 0.26$ & $64.05 \pm 0.32$ & $58.60 \pm 1.10$ & $59.67 \pm 0.27$ & $60.97 \pm 0.48$ \\
 pottedplant & $38.29 \pm 1.23$ & $11.55 \pm 0.71$ & $31.42 \pm 0.31$ & $28.09 \pm 0.24$ & $24.91 \pm 0.12$ & $24.16 \pm 0.51$ \\
 road        & $39.84 \pm 0.64$ & $18.67 \pm 1.09$ & $29.96 \pm 2.20$ & $26.56 \pm 2.51$ & $25.48 \pm 1.25$ & $29.18 \pm 1.40$ \\
 sheep       & $54.34 \pm 0.98$ & $10.72 \pm 0.09$ & $32.62 \pm 3.89$ & $27.46 \pm 0.46$ & $30.55 \pm 1.20$ & $32.02 \pm 2.90$ \\
 sky         & $87.78 \pm 0.40$ & $73.27 \pm 0.07$ & $71.35 \pm 2.91$ & $71.74 \pm 1.09$ & $77.42 \pm 0.61$ & $71.55 \pm 0.21$ \\
 sofa        & $33.56 \pm 2.26$ & $13.95 \pm 1.86$ & $24.26 \pm 1.58$ & $28.81 \pm 0.58$ & $27.07 \pm 1.29$ & $24.34 \pm 1.42$ \\
 track       & $51.97 \pm 0.73$ & $22.59 \pm 1.98$ & $41.96 \pm 1.18$ & $33.81 \pm 1.06$ & $40.67 \pm 2.09$ & $37.02 \pm 0.59$ \\
 train       & $71.39 \pm 1.07$ & $37.66 \pm 1.10$ & $61.06 \pm 1.61$ & $55.68 \pm 0.08$ & $56.54 \pm 1.03$ & $53.58 \pm 0.17$ \\
 tree        & $64.17 \pm 0.84$ & $34.11 \pm 0.44$ & $57.14 \pm 0.71$ & $56.35 \pm 0.68$ & $54.91 \pm 0.01$ & $51.80 \pm 0.56$ \\
 tvmonitor   & $59.29 \pm 1.38$ & $42.13 \pm 0.77$ & $42.23 \pm 0.24$ & $33.31 \pm 1.36$ & $37.28 \pm 0.87$ & $31.34 \pm 0.16$ \\
 wall        & $54.42 \pm 0.43$ & $35.53 \pm 0.15$ & $44.63 \pm 0.81$ & $43.95 \pm 0.86$ & $42.01 \pm 1.45$ & $39.73 \pm 0.67$ \\
 water       & $70.70 \pm 0.90$ & $38.16 \pm 0.99$ & $43.65 \pm 1.11$ & $28.62 \pm 1.16$ & $26.77 \pm 1.33$ & $32.49 \pm 0.45$ \\

\hline
\end{tabular}
\end{table*}

\section{Class-wise analysis}
In the following, we investigate on the Cityscapes dataset the performance differences between the baseline model and the models trained on stylized images which achieved the highest $\text{IoU}_\text{Shape}$ in CDSB. Results are listed in \Cref{tab:class_ious}.
The relative order of class IoUs within model architectures between baseline networks and those trained on stylized data remains mostly identical, with a few exceptions, such as the classes \textit{train}, \textit{bus} and \textit{motorcycle} with lower values. This could be caused by the low frequency of these classes, and bicycles and motorcycles becoming harder to distinguish in the stylized images.
A comparison between the model architectures shows that the models have similar class IoUs in most cases.
Notably, SegFormer models experience a higher drop in IoU for the classes \textit{truck}, \textit{wall} and \textit{rider} than expected from the drop in mIoU. In contrast, \textit{train} is significantly more difficult to segment for the convolutional neural network trained with stylization.
These differences could stem from the different model encoding mechanisms as well as the randomness of the applied style combined with the low frequency at which the class occurs (cf.\ the high variance for the class \textit{train}).

\begin{figure*}[htbp]
    \centering

    % Row 1: Contrast Changes
    \textbf{Contrast changes}\\[1ex]
    \begin{minipage}[t]{\textwidth}
        \centering
        \begin{minipage}[b]{0.14\textwidth}
            \includegraphics[width=\textwidth]{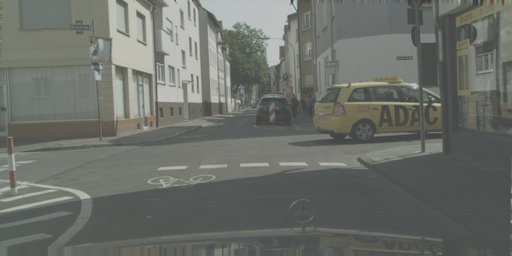}
            \centering\scriptsize 0.5
        \end{minipage}\hfill
        \begin{minipage}[b]{0.14\textwidth}
            \includegraphics[width=\textwidth]{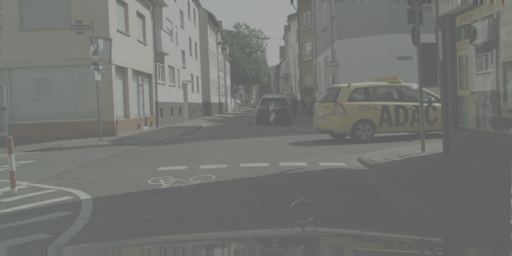}
            \centering\scriptsize 0.3
        \end{minipage}\hfill
        \begin{minipage}[b]{0.14\textwidth}
            \includegraphics[width=\textwidth]{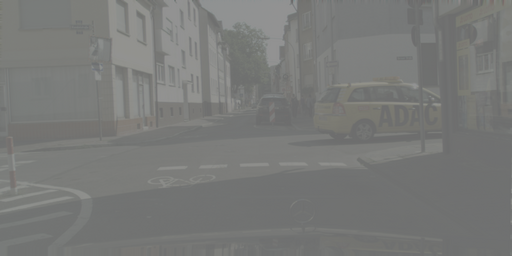}
            \centering\scriptsize 0.15
        \end{minipage}\hfill
        \begin{minipage}[b]{0.14\textwidth}
            \includegraphics[width=\textwidth]{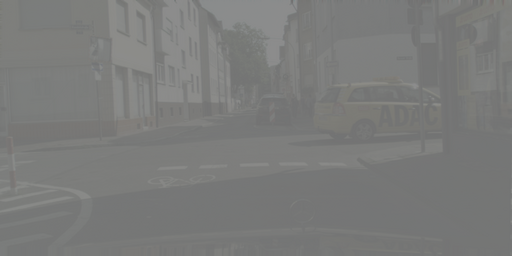}
            \centering\scriptsize 0.1
        \end{minipage}\hfill
        \begin{minipage}[b]{0.14\textwidth}
            \includegraphics[width=\textwidth]{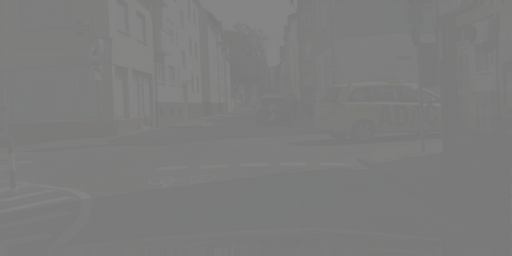}
            \centering\scriptsize 0.05
        \end{minipage}\hfill
        \begin{minipage}[b]{0.14\textwidth}
            \includegraphics[width=\textwidth]{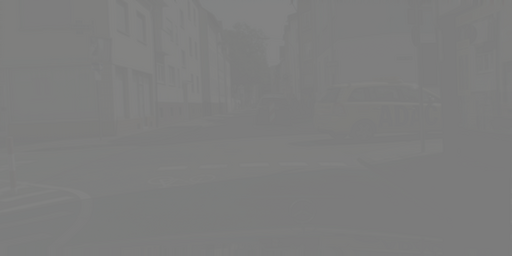}
            \centering\scriptsize 0.03
        \end{minipage}\hfill
        \begin{minipage}[b]{0.14\textwidth}
            \includegraphics[width=\textwidth]{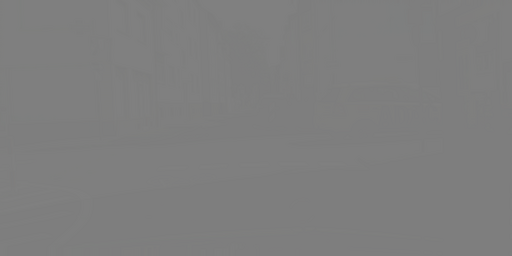}
            \centering\scriptsize 0.01
        \end{minipage}
    \end{minipage}

    \vspace{1pt}

    % Row 2: Uniform Noise
    \textbf{Uniform noise}\\[1ex]
    \begin{minipage}[t]{\textwidth}
        \centering
        \begin{minipage}[b]{0.14\textwidth}
            \includegraphics[width=\textwidth]{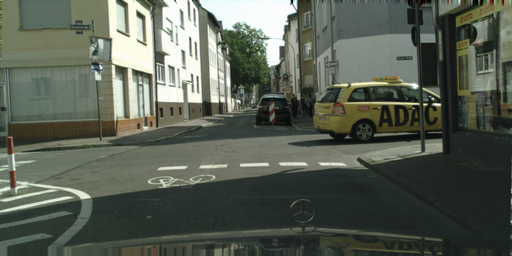}
            \centering\scriptsize 0.03
        \end{minipage}\hfill
        \begin{minipage}[b]{0.14\textwidth}
            \includegraphics[width=\textwidth]{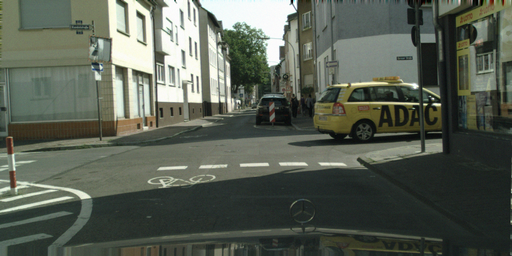}
            \centering\scriptsize 0.05
        \end{minipage}\hfill
        \begin{minipage}[b]{0.14\textwidth}
            \includegraphics[width=\textwidth]{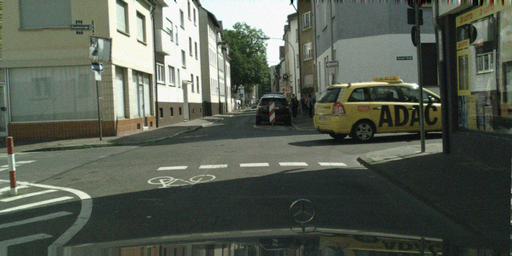}
            \centering\scriptsize 0.1
        \end{minipage}\hfill
        \begin{minipage}[b]{0.14\textwidth}
            \includegraphics[width=\textwidth]{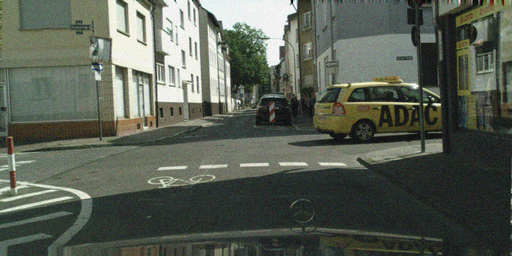}
            \centering\scriptsize 0.2
        \end{minipage}\hfill
        \begin{minipage}[b]{0.14\textwidth}
            \includegraphics[width=\textwidth]{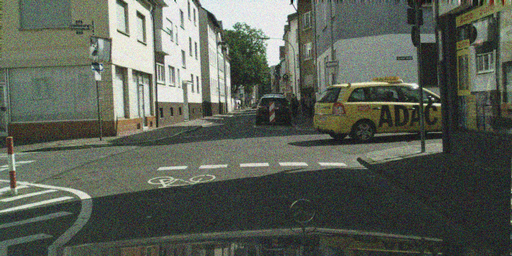}
            \centering\scriptsize 0.35
        \end{minipage}\hfill
        \begin{minipage}[b]{0.14\textwidth}
            \includegraphics[width=\textwidth]{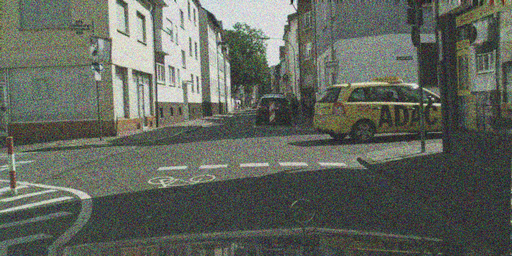}
            \centering\scriptsize 0.6
        \end{minipage}\hfill
        \begin{minipage}[b]{0.14\textwidth}
            \includegraphics[width=\textwidth]{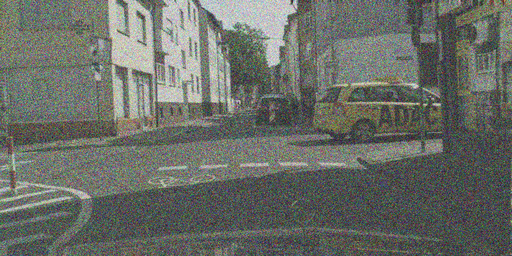}
            \centering\scriptsize 0.9
        \end{minipage}
    \end{minipage}

    \vspace{1pt}

    % Row 3: Low Pass Filter
    \textbf{Low pass filter}\\[1ex]
    \begin{minipage}[t]{\textwidth}
        \centering
        \begin{minipage}[b]{0.14\textwidth}
            \includegraphics[width=\textwidth]{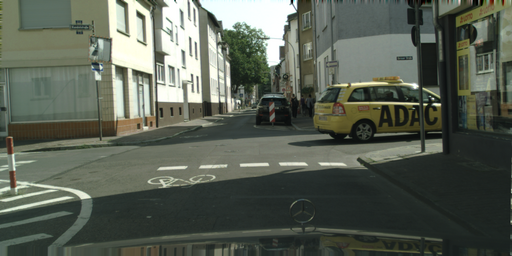}
            \centering\scriptsize 0.0
        \end{minipage}\hfill
        \begin{minipage}[b]{0.14\textwidth}
            \includegraphics[width=\textwidth]{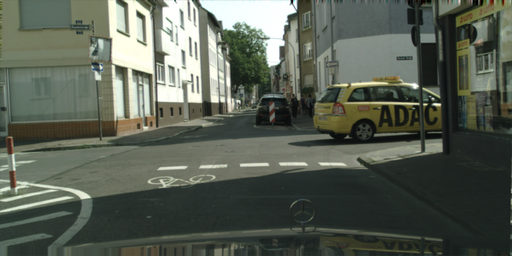}
            \centering\scriptsize 1.0
        \end{minipage}\hfill
        \begin{minipage}[b]{0.14\textwidth}
            \includegraphics[width=\textwidth]{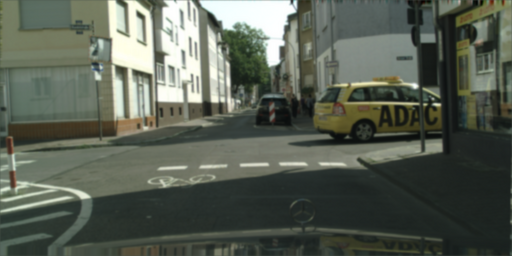}
            \centering\scriptsize 3.0
        \end{minipage}\hfill
        \begin{minipage}[b]{0.14\textwidth}
            \includegraphics[width=\textwidth]{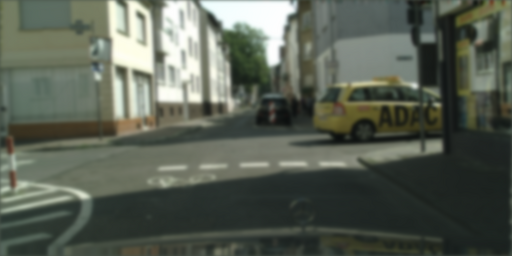}
            \centering\scriptsize 7.0
        \end{minipage}\hfill
        \begin{minipage}[b]{0.14\textwidth}
            \includegraphics[width=\textwidth]{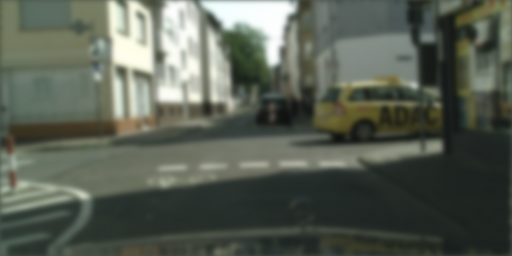}
            \centering\scriptsize 10.0
        \end{minipage}\hfill
        \begin{minipage}[b]{0.14\textwidth}
            \includegraphics[width=\textwidth]{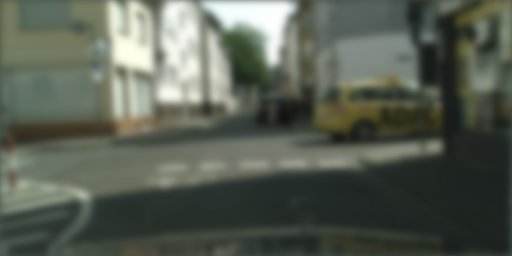}
            \centering\scriptsize 15.0
        \end{minipage}\hfill
        \begin{minipage}[b]{0.14\textwidth}
            \includegraphics[width=\textwidth]{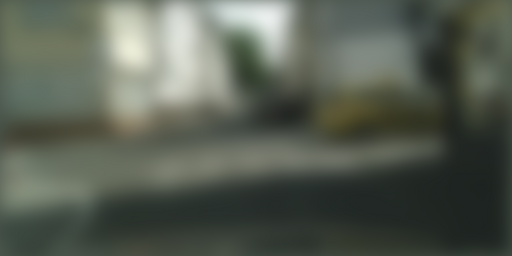}
            \centering\scriptsize 40.0
        \end{minipage}
    \end{minipage}

    \vspace{1pt}

    % Row 4: High Pass Filter
    \textbf{High pass filter}\\[1ex]
    \begin{minipage}[t]{\textwidth}
        \centering
        \begin{minipage}[b]{0.14\textwidth}
            \includegraphics[width=\textwidth]{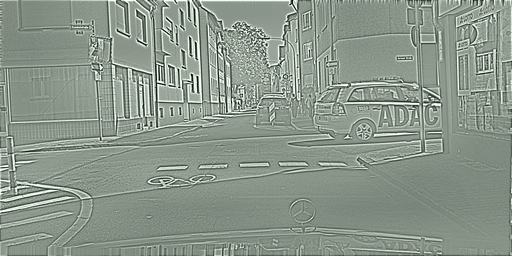}
            \centering\scriptsize 3.0
        \end{minipage}\hfill
        \begin{minipage}[b]{0.14\textwidth}
            \includegraphics[width=\textwidth]{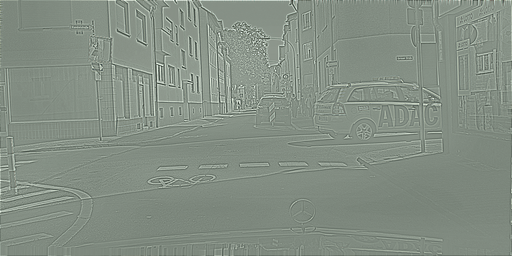}
            \centering\scriptsize 1.5
        \end{minipage}\hfill
        \begin{minipage}[b]{0.14\textwidth}
            \includegraphics[width=\textwidth]{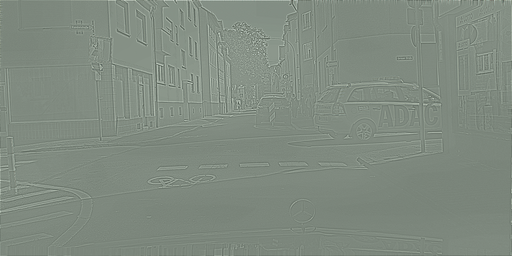}
            \centering\scriptsize 1.0
        \end{minipage}\hfill
        \begin{minipage}[b]{0.14\textwidth}
            \includegraphics[width=\textwidth]{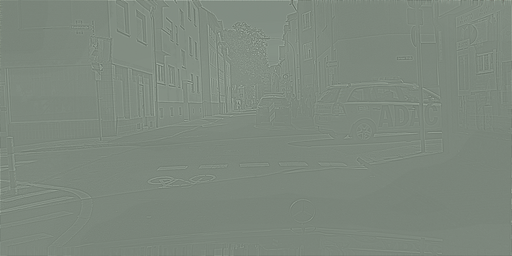}
            \centering\scriptsize 0.7
        \end{minipage}\hfill
        \begin{minipage}[b]{0.14\textwidth}
            \includegraphics[width=\textwidth]{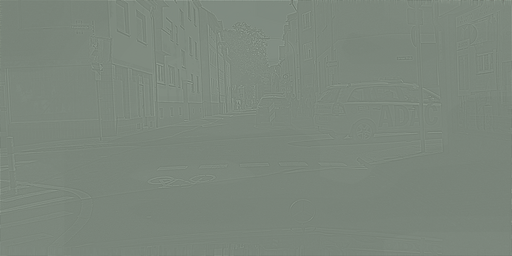}
            \centering\scriptsize 0.55
        \end{minipage}\hfill
        \begin{minipage}[b]{0.14\textwidth}
            \includegraphics[width=\textwidth]{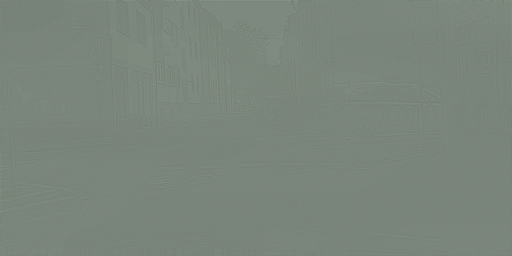}
            \centering\scriptsize 0.45
        \end{minipage}\hfill
        \begin{minipage}[b]{0.14\textwidth}
            \includegraphics[width=\textwidth]{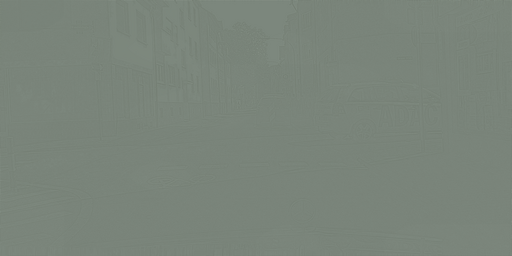}
            \centering\scriptsize 0.4
        \end{minipage}
    \end{minipage}

    \vspace{1pt}

    % Row 5: Phase Noise
    \textbf{Phase noise}\\[1ex]
    \begin{minipage}[t]{\textwidth}
        \centering
        \begin{minipage}[b]{0.14\textwidth}
            \includegraphics[width=\textwidth]{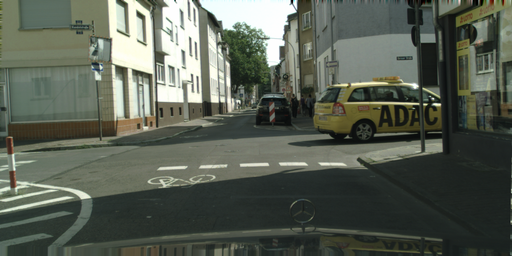}
            \centering\scriptsize 0°
        \end{minipage}\hfill
        \begin{minipage}[b]{0.14\textwidth}
            \includegraphics[width=\textwidth]{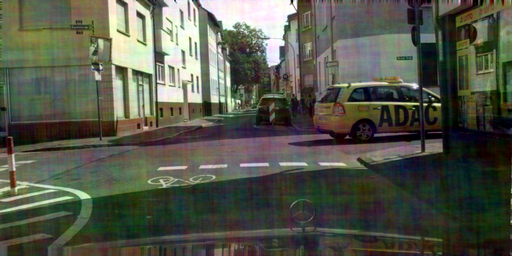}
            \centering\scriptsize 30°
        \end{minipage}\hfill
        \begin{minipage}[b]{0.14\textwidth}
            \includegraphics[width=\textwidth]{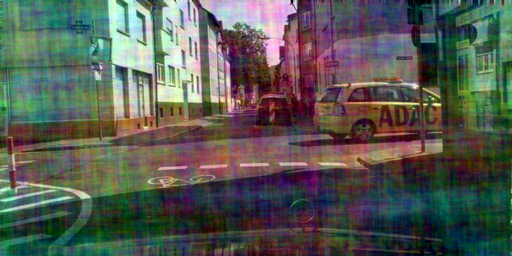}
            \centering\scriptsize 60°
        \end{minipage}\hfill
        \begin{minipage}[b]{0.14\textwidth}
            \includegraphics[width=\textwidth]{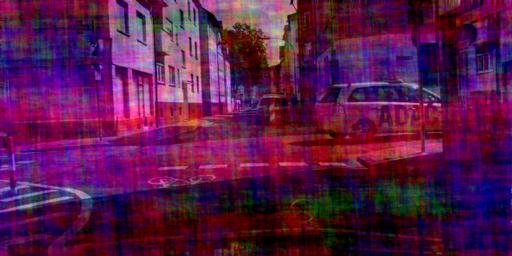}
            \centering\scriptsize 90°
        \end{minipage}\hfill
        \begin{minipage}[b]{0.14\textwidth}
            \includegraphics[width=\textwidth]{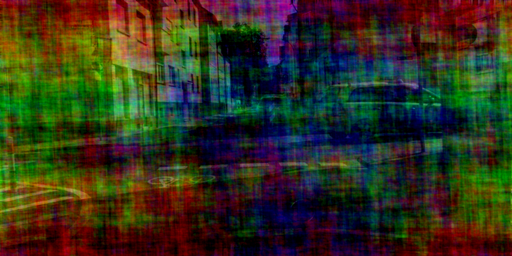}
            \centering\scriptsize 120°
        \end{minipage}\hfill
        \begin{minipage}[b]{0.14\textwidth}
            \includegraphics[width=\textwidth]{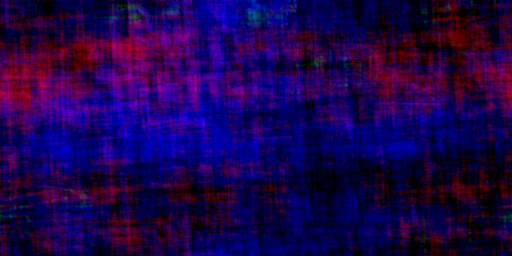}
            \centering\scriptsize 150°
        \end{minipage}\hfill
        \begin{minipage}[b]{0.14\textwidth}
            \includegraphics[width=\textwidth]{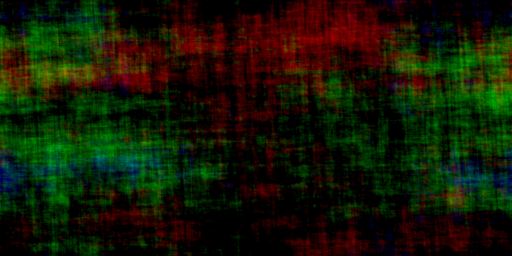}
            \centering\scriptsize 180°
        \end{minipage}
    \end{minipage}

    \caption[Image corruptions visualization]{Illustration of image corruptions with increasing intensity parameters. Zero parameters correspond to the original image. Best viewed with zoom.}
    \label{fig:distortion_visualization}
\end{figure*}

For the PASCAL Context dataset, see \Cref{tab:class_miou_pc}, some classes seem to suffer more than other when stylization is applied. As discussed in the main paper, the class \textit{sheep} looses roughly $20$ percent points. This suggests that segmenting a sheep relies strongly on the texture and it is hard to learn their shape features due to texture augmentation. 
To investigate semantic groups, we split the classes into foreground and background classes to analyze whether foreground objects benefit more or less from stylization. We define following classes as foreground: 
\textit{aeroplane},
\textit{bicycle},
\textit{bird},
\textit{boat},
\textit{bottle},
\textit{bus},
\textit{car}
\textit{cat},
\textit{chair},
\textit{cow},
\textit{diningtable},
\textit{dog},
\textit{horse},
\textit{keyboard},
\textit{motorbike},
\textit{person},
\textit{pottedplant},
\textit{sheep},
\textit{sofa},
\textit{train},
\textit{tvmonitor}.
The remaining classes are defined as background.
We first observe that the averaged performance of the foreground classes when evaluated on the original data are on par with the averaged performance of the background classes for all trained models (baseline, Voronoi 1, 4, 8 and 16).
With respect to the shape bias, we observe an increase in the shape component for an increasing amount of Voronoi cells from 1 to 8. However, the texture component increases simultaneously which in total leads to a drop of the shape bias as observed on mIoU level. For the averaged shape bias over the background classes we observe the contrary behavior for the shape component, i.e., $\text{IoU}_\text{Shape}$ decreases when increasing the amount of differently stylized Voronoi cells. This leads to a generally lower shape bias for the background classes compared to the foreground classes by comparable mIoU performance on the original dataset. Nevertheless, stylization improves the shape biases compared to the baseline for all classes and varying amount of Voronoi cells.

\section{Image corruptions for robustness experiments}\label{app:sec:corruptions}
To generate the distorted images, the script to apply image corruptions from \cite{geirhos2020generalisationhumansdeepneural} was modified to distort RGB images rather than grayscale images by applying corruptions per channel. 

The individual corruptions are implemented as follows:
Uniform noise $N$ adds entry-wise uniform random noise to the input image \( I \in [0,1]^{H \times W \times 3} \):
\begin{equation}
    I_{\text{noise}} = I + N
\end{equation}
Here, every entry of $N$ is uniformly distributed in $(-\eta,\eta)$ for $ \eta \in \{0.0, 0.03, 0.05, 0.1, 0.2, 0.35, 0.6, 0.9\}$.

Contrast reduction scales image contrast toward the midpoint ($0.5$):
\begin{equation}
    I_{\text{contrast}} = \frac{1 - c}{2} + c \cdot I,
\end{equation}
where the scaling factor \( c \in \{1,3,5,10,15,30,50,100\}\% \) determines the contrast level.

Low-pass filtering applies a Gaussian blur, removing high-frequency image details:
\begin{equation}
\begin{aligned}
    I_{\text{low-pass}}(x,y,c) &= (I_c * G_{\sigma})(x,y), \\ G_{\sigma}(x,y)&=\frac{1}{2\pi\sigma^2} e^{-\frac{x^2+y^2}{2\sigma^2}}
\end{aligned}
\end{equation}
with standard deviations \( \sigma \in \{1, 3, 7, 10, 15, 40\} \) and a kernel truncated at \(4\sigma\).

High-pass filtering subtracts the Gaussian-blurred image (low-pass) from the original, enhancing edges and fine detail:
\begin{equation}
\begin{aligned}
    I_{\text{high-pass}}(x,y,c) = I_c&(x,y) - (I_c * G_{\sigma})(x,y) \\ &+ (\mu_c^{\text{dataset}} - \mu_c^{I_c(x,y) - (I_c * G_{\sigma})(x,y)})
\end{aligned}
\end{equation}
where \(\mu_c^{\text{dataset}}\approx(0.2945, 0.3334, 0.2949)\) are the mean channel intensities of the Cityscapes dataset, \(\mu_c^{\text{high-pass}}\) is the mean intensity of the high-pass filtered image channel \(c\), \(\sigma \in \{0.4, 0.45, 0.55, 0.7, 1, 1.5, 3\}\) controls the filter's spatial scale, and the Gaussian kernel is truncated at \(4\sigma\).

Phase noise randomizes the Fourier-domain phase of image frequencies, disturbing structural coherence without changing the amplitude spectrum. For the Fourier-transform $\mathcal{F}$, spatial frequencies in horizontal and vertical direction $u, v$ and the imaginary unity $j$:
\begin{equation}
    \mathcal{F}\{I_{\text{phase}}\}(u,v) = |\mathcal{F}\{I\}(u,v)| \cdot e^{j(\angle \mathcal{F}\{I\}(u,v) + \phi(u,v))},
\end{equation}
where \(\phi(u,v)\) is uniformly distributed random noise in radians, with width \(w \in \{0, 30, 60, 90, 120, 150, 180\}\) degrees.

After each corruption operation, pixel values are clipped to the range $[0,1]$ for floating-point images. The Gaussian filter for high- and low-pass filtering and the Fast Fourier Transform (FFT) utilize the implementations of SciPy.
\Cref{fig:distortion_visualization} shows the described corruptions applied to an example image of the Cityscapes dataset with increasing intensity from left to right.

\section{Detailed image corruption robustness}

\begin{figure*}[htbp]
    \centering

    % Row 1 – Contrast
    \begin{minipage}[b]{0.495\textwidth}
        \includegraphics[width=\textwidth]{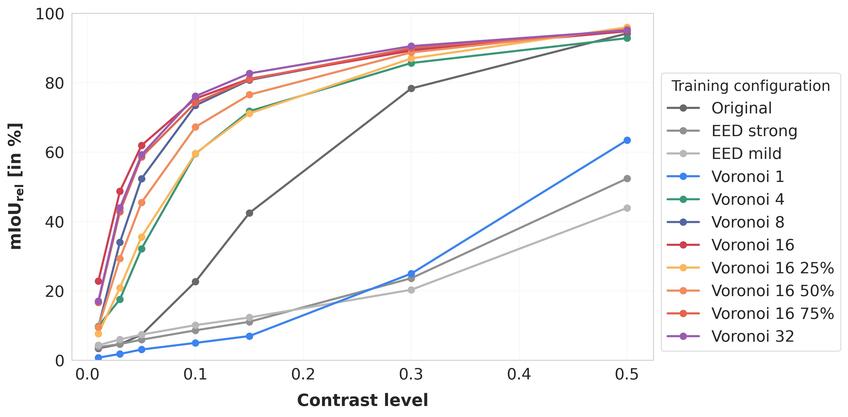}
        \centering\scriptsize DeepLabV3+: Contrast
    \end{minipage}\hfill
    \begin{minipage}[b]{0.495\textwidth}
        \includegraphics[width=\textwidth]{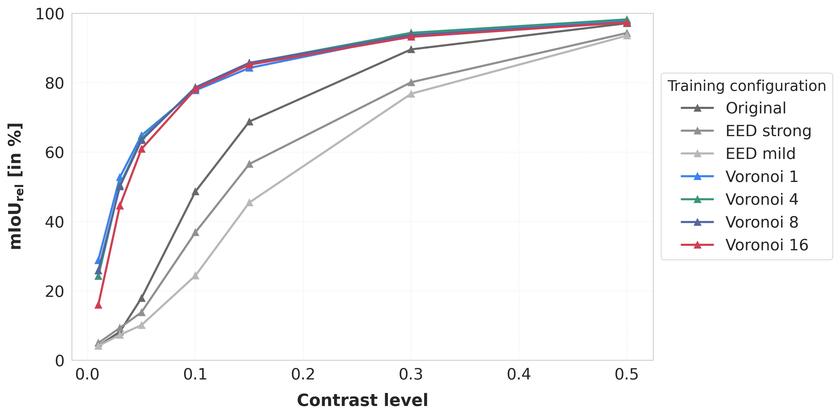}
        \centering\scriptsize SegFormer: Contrast
    \end{minipage}

    \vspace{0.25ex}\par

    % Row 2 – Low-pass Filter
    \begin{minipage}[b]{0.492\textwidth}
        \includegraphics[width=\textwidth]{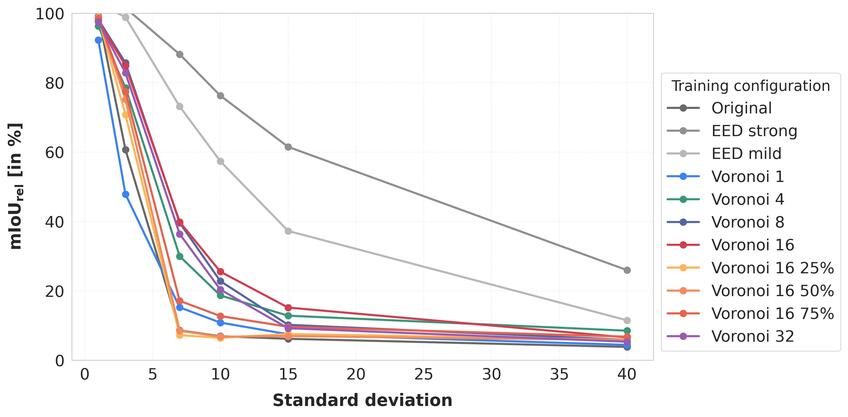}
        \centering\scriptsize DeepLabV3+: Low pass filter
    \end{minipage}\hfill
    \begin{minipage}[b]{0.492\textwidth}
        \includegraphics[width=\textwidth]{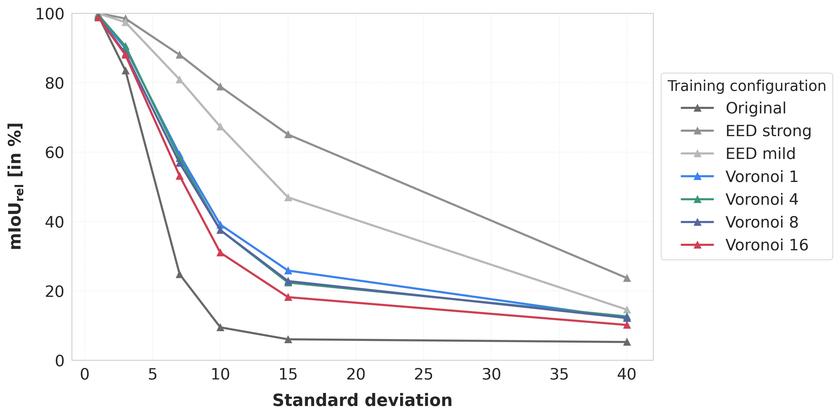}
        \centering\scriptsize SegFormer: Low pass filter
    \end{minipage}

    \vspace{0.25ex}\par

    % Row 3 – High-pass Filter
    \begin{minipage}[b]{0.492\textwidth}
        \includegraphics[width=\textwidth]{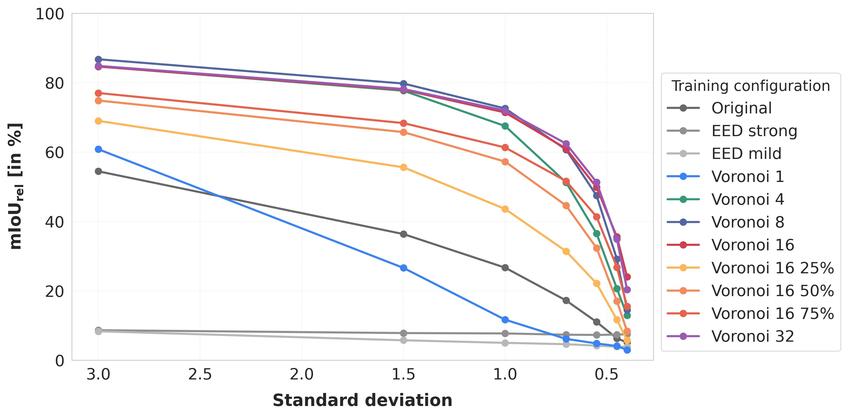}
        \centering\scriptsize DeepLabV3+: High pass filter
    \end{minipage}\hfill
    \begin{minipage}[b]{0.492\textwidth}
        \includegraphics[width=\textwidth]{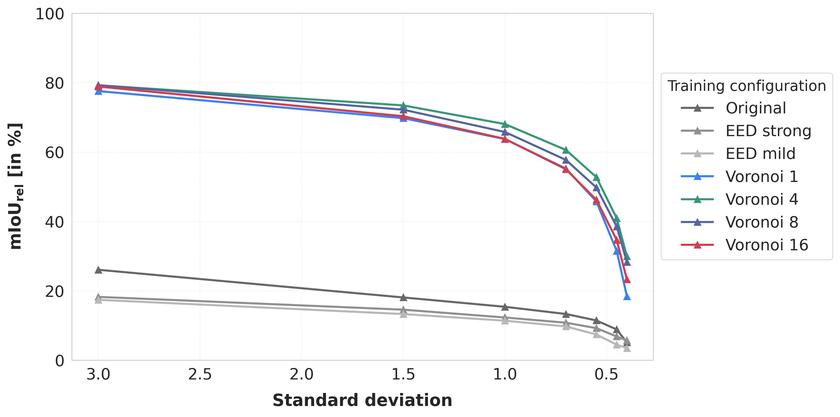}
        \centering\scriptsize SegFormer: High pass filter
    \end{minipage}

    \vspace{0.25ex}\par

    % Row 4 – Phase Noise
    \begin{minipage}[b]{0.492\textwidth}
        \includegraphics[width=\textwidth]{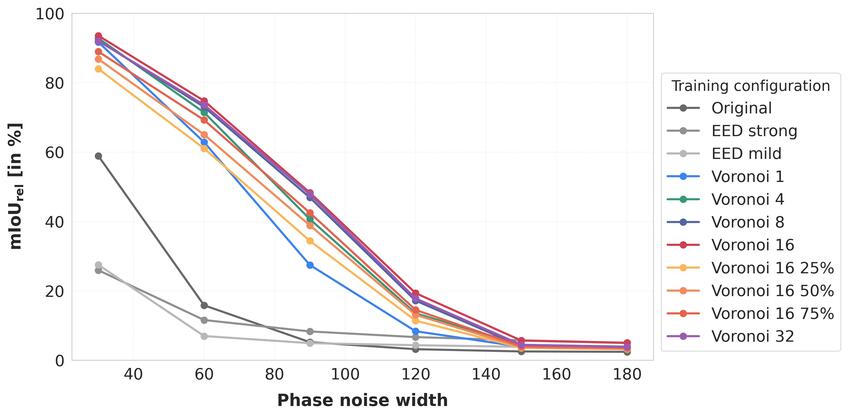}
        \centering\scriptsize DeepLabV3+: Phase noise
    \end{minipage}\hfill
    \begin{minipage}[b]{0.492\textwidth}
        \includegraphics[width=\textwidth]{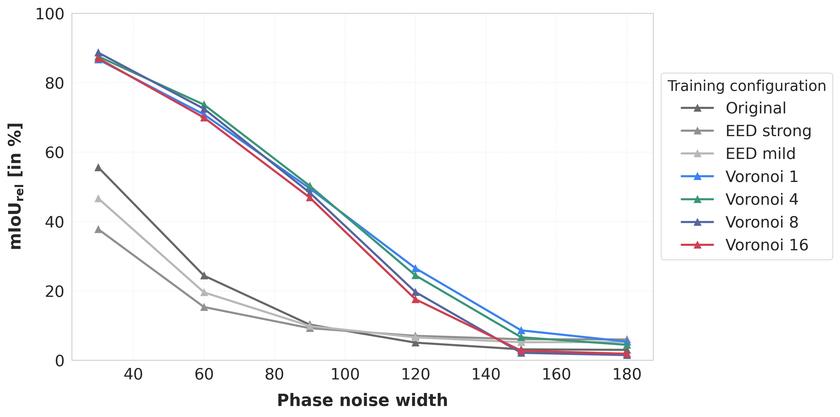}
        \centering\scriptsize SegFormer: Phase noise
    \end{minipage}

    \vspace{0.25ex}\par

    % Row 5 – Uniform Noise
    \begin{minipage}[b]{0.492\textwidth}
        \includegraphics[width=\textwidth]{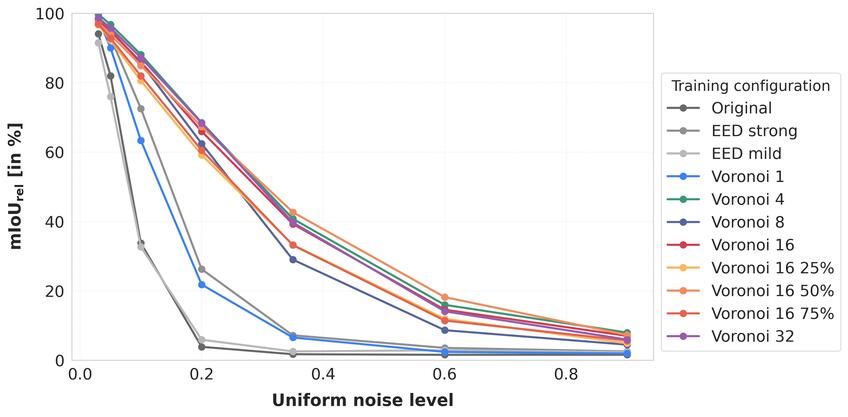}
        \centering\scriptsize DeepLabV3+: Uniform noise
    \end{minipage}\hfill
    \begin{minipage}[b]{0.492\textwidth}
        \includegraphics[width=\textwidth]{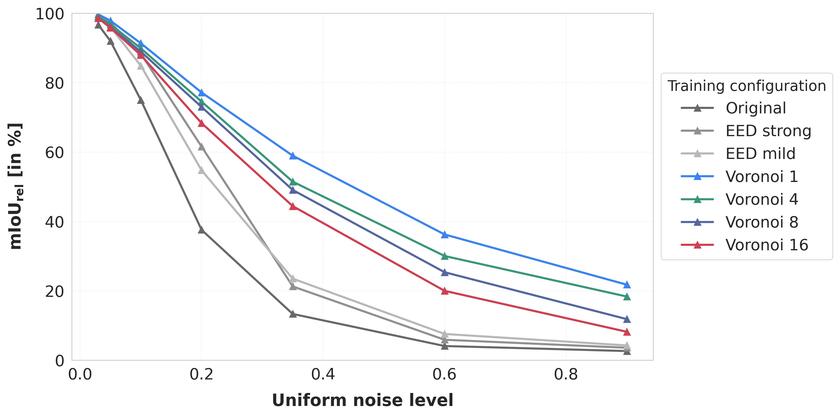}
        \centering\scriptsize SegFormer: Uniform noise
    \end{minipage}
    \vspace{5pt}
    \caption[Distortions detailed Analysis]{Comparison of different distortions on DeepLabV3+ (left) and SegFormer (right). Scores are normalized 
    by mIoU on undistorted images.}
    \label{fig:distortion_comparison}
\end{figure*}

Here, we analyze the impact of image corruptions on the mIoU of semantic segmentation models on a per-corruption-type level. For each type, the impact of different corruption levels is visualized in \cref{fig:distortion_comparison}.

\paragraph{Contrast.}We observe that all models maintain near‑baseline accuracy down to approximately $15\%$ contrast, with mIoU falling off sharply below that threshold. The EED baselines and the Voronoi 1 DeepLabV3+ variant begin to degrade earlier, whereas convolutional networks trained on stylized images with $n>1$ patches sustain high performance even at very low contrasts. SegFormer exhibits the greatest resilience: its stylized‑training variants retain over $20\%$ of their original mIoU at just $1\%$ contrast, with only the model trained on stylized images with $n=16$ Voronoi patches falling below this level.
\paragraph{Low-pass filter.}

\begin{figure*}[htbp]
    \centering

    %--- DeepLabV3+ untargeted FGSM ---
    \begin{minipage}[b]{0.45\textwidth}
        \centering
        \includegraphics[width=\textwidth]{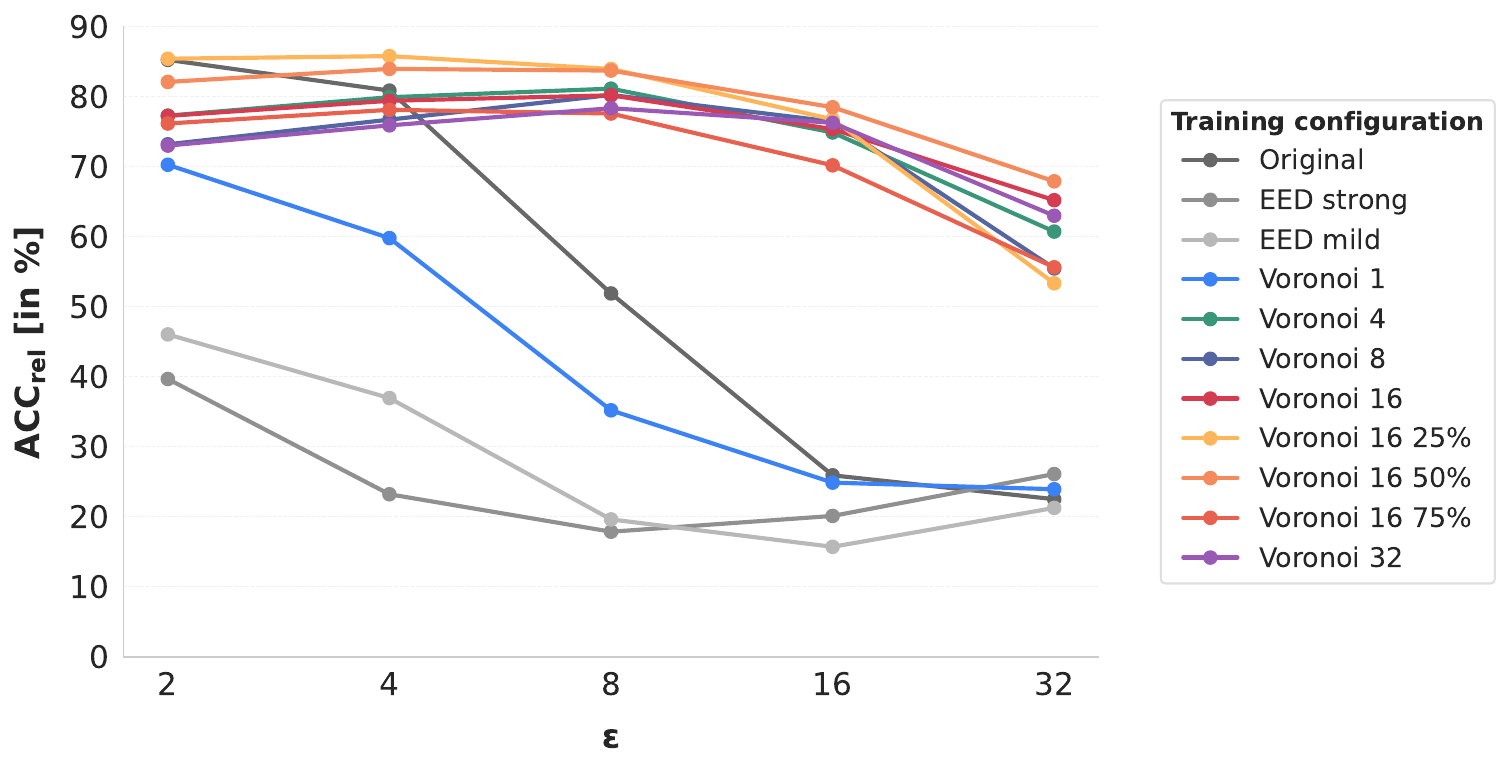}
        {\scriptsize DeepLabV3+}
    \end{minipage}%\hfill
    %--- SegFormer untargeted FGSM ---
    \begin{minipage}[b]{0.45\textwidth}
        \centering
        \includegraphics[width=\textwidth]{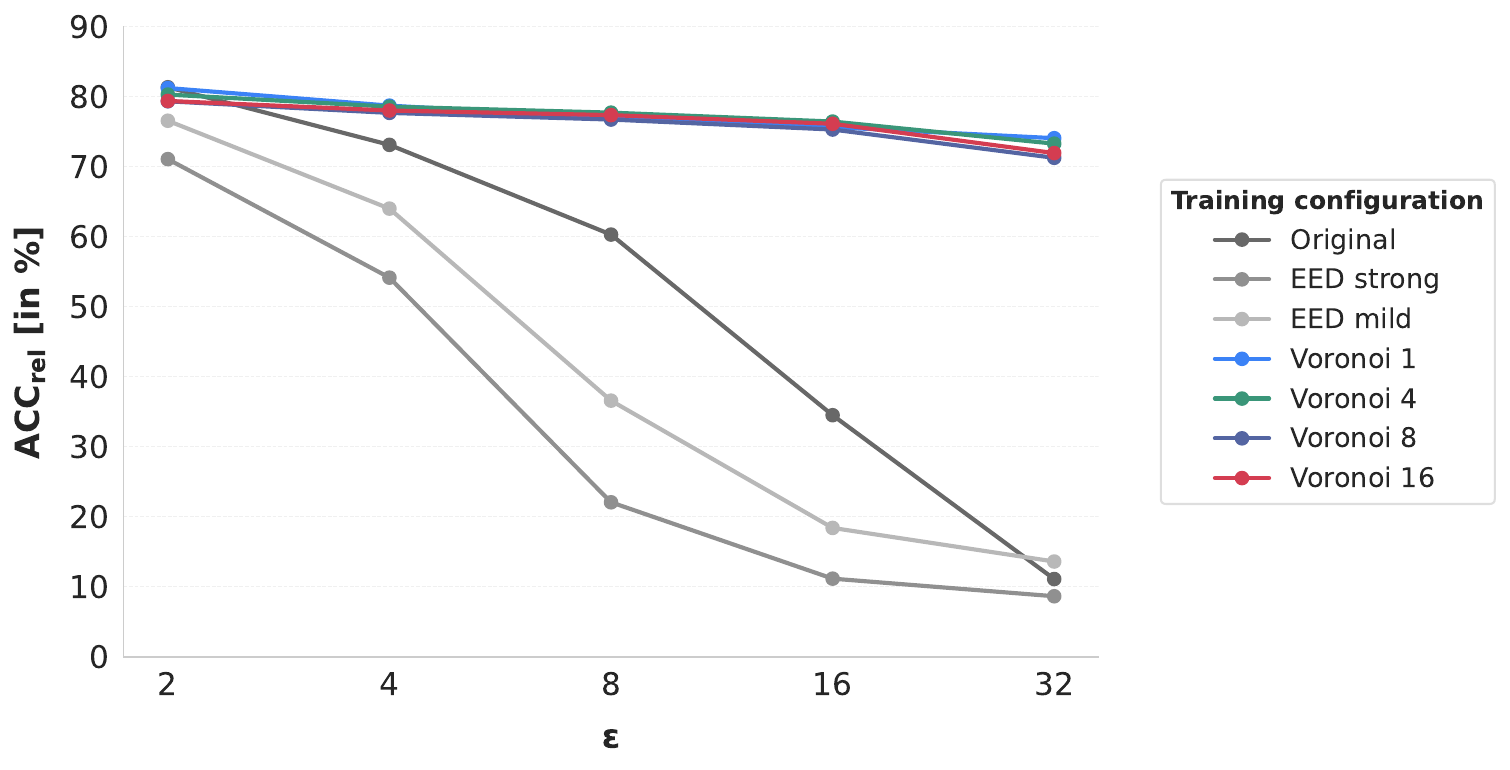}
        {\scriptsize SegFormer}
    \end{minipage}

    \caption[Robustness against untargeted FGSM attacks under variation of $\varepsilon$]{Visualization of robustness to untargeted FGSM attacks with varying $\varepsilon$, measured by relative accuracy. The results show mean values for all three runs per training configuration.}
    \label{fig:fgsm_untargeted_progression}
\end{figure*}

\begin{figure*}[htbp]
    \centering

    %--- DeepLabV3+ plot ---
    \begin{minipage}[b]{0.45\textwidth}
        \centering
        \includegraphics[width=\textwidth]{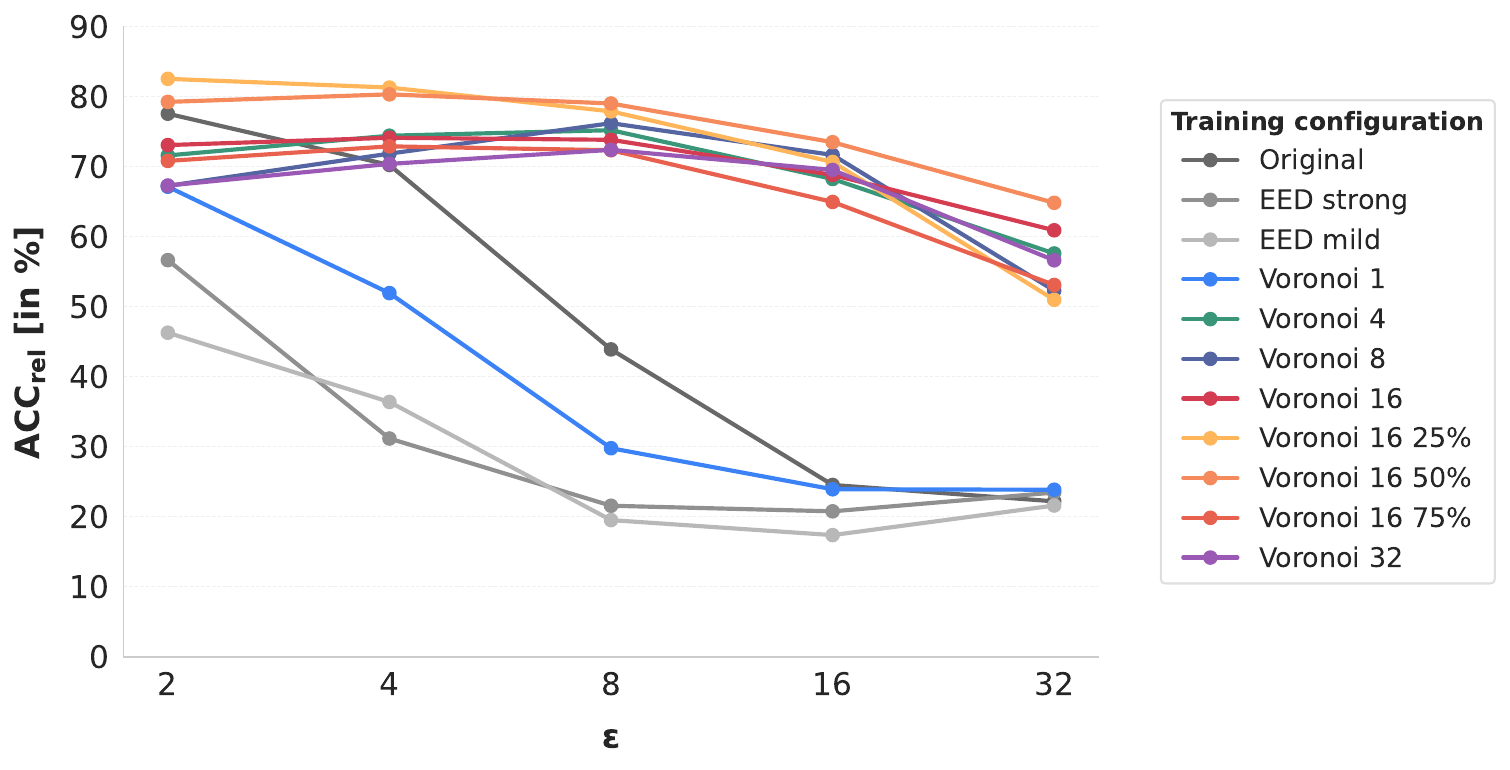}
        {\scriptsize DeepLabV3+}
    \end{minipage}%\hfill
    %--- SegFormer plot ---
    \begin{minipage}[b]{0.45\textwidth}
        \centering
        \includegraphics[width=\textwidth]{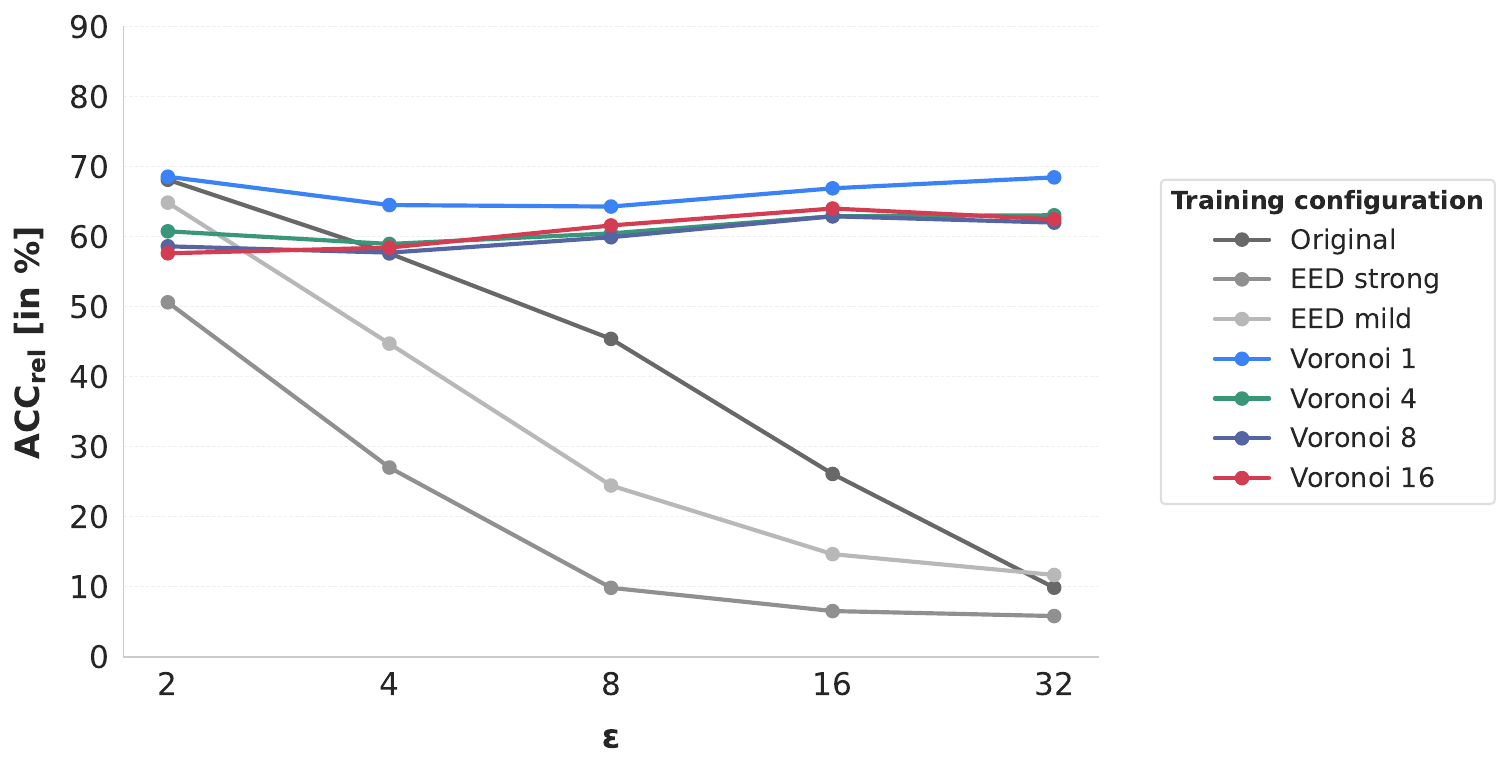}
        {\scriptsize SegFormer}
    \end{minipage}

    \caption[Robustness against targeted FGSM Attacks under Variation of $\varepsilon$]{Visualization of robustness to targeted FGSM attacks with varying $\varepsilon$, measured by relative accuracy. The results show mean values for all three runs per training configuration.}
    \label{fig:fgsm_targeted_progression}
\end{figure*}

We see the strongest results achieved by the EED baseline networks, as expected by the construction of edge-enhancing diffusion which applies Gaussian (low-pass) filters to blur textures. In the case of the DeepLabV3+, only fully patch-wise stylized training images appear to increase the robustness whereas partially stylized training images show almost identical results to the original image baseline models. The Voronoi 16 models show a slightly better performance up to a standard deviation of $15$ percent points. The SegFormer models show a bit more robustness to the Gaussian blur with a slower decay at the lower distortion intensities and the best results for Voronoi 1 models.

\paragraph{High-pass filter.}
The Voronoi 4, 8, 16 and 32 DeepLabV3+ networks show highly increased robustness to high-pass filtering in comparison to other training configurations. The Voronoi 16 model shows the least decay at the highest distortion intensity. The networks trained on partially stylized images appear to be more robust than the baseline network with progressions between the baseline network and networks trained on fully stylized images. A higher proportion of stylized segments also corresponds to higher relative mIoUs in the experiment. The SegFormer baseline model appears to be a lot less robust to high-pass filtering than the DeepLabV3+ baseline model which can be greatly improved by stylizing the training images. The Voronoi 4 model, shows the highest improvement in robustness. In general, the networks trained on stylized data show a similar progression. 
\paragraph{Phase noise.}
Both network architectures display similar performance decays when faced with noise in the phase space with the SegFormer baseline model dropping less steeply than the DeepLabV3+ baseline model. All networks trained on style-transferred data show increased robustness to phase noise with the convolutional networks trained on images with 8 to 32 fully stylized segments posting the highest relative mIoUs for increasing noise widths. The Voronoi 1, 4 and partially stylized Voronoi 16 variants also show significantly increased robustness in comparison to baseline models.
The SegFormer models present a similar decay to the DeepLabV3+ models trained on segment-wise style-transferred images with the baseline model initially decaying more slowly. The single-style variant shows slightly improved results at the highest noise widths when compared to the variants Voronoi 4 to 16. 
\paragraph{Uniform noise.}
Uniform noise distortion adds random values up to the noise intensity to each pixel and color channel. We observe that the SegFormer models are a lot more robust with slower decays in comparison to the DeepLabV3+. The style-transfer-based training augmentations yield large improvements in robustness for both architectures. Networks trained on partially stylized segments present highly elevated robustness to uniform noise with all tested proportions leading to results similar to those of fully stylized variants. This corresponds to the observations of increased robustness to adversarial attacks observed in the main manuscript. In the case of SegFormer models, the Voronoi 1 variant shows the largest improvements. Compared to the CNN, they are a lot more robust to the introduction of random uniform noise.

\section{Robustness to adversarial attacks}

The model performance for untargeted attacks is displayed in \Cref{fig:fgsm_untargeted_progression}. The DeepLabV3+ baseline model shows a faster decline in relative accuracy when compared to the SegFormer baseline model. Both EED-trained and the Voronoi 1 DeepLabV3+ model show an even faster decline with large drops already at the lowest perturbation strength of $\varepsilon=2$. The SegFormer EED baseline models similarly perform below the original image baseline network. The convolutional network appears to benefit greatly from training data enriched with segment-wise style transfer as seen by the enhanced robustness up to the highest parameter settings. In general, the segment-wise style transfer appears to have positive effects on the model robustness to the tested attack. The partially stylized versions improve the model robustness to the tested attack to an even greater extent. Models trained on $25\%$ or $50\%$ stylized segments even achieve the highest relative accuracies observed in this experiment. The $75\%$ variant unexpectedly performs lower than its 25\% and 50\% counterparts. In summary, the tested convolutional network architecture obtains greatly improved robustness to the tested untargeted FGSM attack by training on stylized segments. 
The SegFormer models show natively higher robustness in all variants, with any model trained on stylized data becoming practically invariant to the adversarial attack perturbation strength after an initial drop to approximately $80\%$ of their Cityscapes validation accuracy. As seen in \Cref{fig:fgsm_untargeted_progression}, larger image perturbation strengths appear to cause almost no drop in relative accuracy.

The targeted adversarial attack (see \Cref{fig:fgsm_targeted_progression}) elicits largely similar responses from the tested models, with DeepLabV3+ models showing almost the same progression and the relative order between networks. The SegFormer networks appear to be more vulnerable to the targeted attack, as seen by the lower relative accuracy of around $70\%$ for the most robust Voronoi 1 network. This indicates a higher success rate for the attack pixel. In the case of SegFormer models, differences begin to emerge, with the Voronoi 1 variant posting the highest robustness and methods with more segments showing more performance degradation. Surprisingly, the relative accuracy increases with larger perturbations, which is an indicator for possible overshooting. The original image SegFormer model achieves results close to the Voronoi 1 SegFormer model for $\varepsilon=2$ before experiencing a steep drop with increasing $\varepsilon$. 

\end{document}